\documentclass{article}
\PassOptionsToPackage{numbers, compress}{natbib}

\usepackage[final,main]{neurips_2026}
\usepackage{adjustbox}
\usepackage{svg}
\usepackage{array}
\newcolumntype{B}[1]{>{\bfseries}#1}

\usepackage[utf8]{inputenc} 
\usepackage[T1]{fontenc}    

\usepackage[dvipsnames,table]{xcolor}
\definecolor{citecolor}{HTML}{0071BC}
\definecolor{linkcolor}{HTML}{ED1C24}
\definecolor{revisionblue}{RGB}{0,90,180}
\newif\ifshowrevisions
\showrevisionstrue
\ifshowrevisions
  \newcommand{\rev}[1]{{\color{revisionblue}#1}}
  
\else
  \newcommand{\rev}[1]{#1}
  
\fi
\usepackage[pagebackref=true,breaklinks=true,colorlinks,bookmarks=false,citecolor=citecolor,linkcolor=linkcolor,urlcolor=gray]{hyperref}

\usepackage{url}            
\usepackage{booktabs}       
\usepackage{amsfonts, amsmath, amssymb, amsthm}       
\usepackage{nicefrac}       
\usepackage{microtype}      
\usepackage{amsthm}
\newtheorem{statement}{Statement}[section]

\usepackage{graphicx}
\usepackage{makecell}
\usepackage[capitalise,noabbrev]{cleveref}
\usepackage{algorithm}
\usepackage{algpseudocode}
\usepackage{amssymb}
\usepackage{subcaption}
\usepackage{diagbox}
\usepackage{wrapfig}
\usepackage{pifont}
\usepackage{enumitem}
\usepackage{hyperref}
\usepackage{url}
\usepackage[font=small]{caption}

\usepackage{caption}
\usepackage{listings}
\usepackage{xcolor}
\usepackage{algorithm}
\usepackage{courier}
\usepackage{color}
\usepackage[british, american]{babel}
\definecolor{mygreen}{rgb}{0,0.6,0}
\definecolor{mygray}{rgb}{0.5,0.5,0.5}
\definecolor{mymauve}{rgb}{0.58,0,0.82}
\definecolor{codefunc}{rgb}{0.85, 0.18, 0.50}
\usepackage{soul}
\newcommand{\hh}{\hspace{-0.15em}}

\theoremstyle{plain}

\theoremstyle{definition}

\theoremstyle{remark}

\usepackage[textsize=tiny]{todonotes}

\newcommand{\hhh}{\hspace{-1em}}

\usepackage{titlesec}
\title{Equilibrium Matching: Generative Modeling with Implicit Energy-Based Models}

\author{%
  Runqian Wang \\
  MIT \\
  \And
  Yilun Du \\
  Harvard University \\
}

\begin{document}

\maketitle

\vspace{-20pt}
\begin{abstract}
  
We introduce \textit{Equilibrium Matching} (EqM), a generative modeling framework built from an equilibrium dynamics perspective. EqM discards the non-equilibrium, time-conditional dynamics in traditional diffusion and flow-based generative models and instead learns the equilibrium gradient of an implicit energy landscape. Through this approach, we can adopt an optimization-based sampling process at inference time, where samples are obtained by gradient descent on the learned landscape with adjustable step sizes, adaptive optimizers, and adaptive compute. {Empirically, EqM achieves the best FID among the tested methods, reaching 1.90 on ImageNet 256$\times$256, while remaining competitive on other metrics. Theoretically, we show under standard assumptions that real images are attracting critical points of the learned field and the energy interpretation applies when the field is conservative.} Beyond generation, EqM is a flexible framework that naturally handles tasks including partially noised image denoising, OOD detection, and image composition. By replacing time-conditional velocities with a unified equilibrium landscape, EqM offers a tighter bridge between flow and energy-based models and a simple route to optimization-driven inference.
\end{abstract}

\section{Introduction}
  
Generative modeling has advanced rapidly with diffusion and flow-based methods \citep{sohl,ddpm,ddim,fm,recflow}, which map simple noise distributions to complex data by defining a forward noising process and learning its reverse. While they achieve state-of-the-art sample quality \citep{nichol2021improved,dhariwal2021diffusion,edm}, these models employ non-equilibrium dynamics at both training and inference. Diffusion/flow models are conditioned on the input timestep and learn distinct dynamics for inputs at different noise levels. This imposes practical constraints such as a noise-level schedule and a fixed integration horizon.

Existing approaches that attempt to learn equilibrium dynamics suffer from different problems. Sun et al.~\cite{sun2025noise} have shown that forcing diffusion models to learn equilibrium dynamics by simply removing time (noise\footnote{In the context of this work, time conditioning and noise conditioning are equivalent.}) conditioning leads to worse generation quality. Energy-based models (EBMs) \citep{lecun2006tutorial, du2019implicit} directly learn equilibrium energy landscapes, but they often suffer from training instabilities \citep{du2020improved} and poor sample quality \citep{du2019implicit, nijkamp2020anatomy}. More recent approaches such as Energy Matching \citep{balcerak2025energy} involve separate training stages and are difficult to scale on large-scale datasets.

In this work, we introduce \textit{Equilibrium Matching} (EqM), a generative modeling framework from an equilibrium perspective. EqM replaces the time-conditional non-equilibrium dynamics of diffusion/flow models with a single time-invariant equilibrium gradient over an implicit energy landscape. We hypothesize that the quality degradation in noise-unconditional diffusion models originates from the incompatibility between the target gradient and equilibrium dynamics. To this end, we introduce a new family of target gradients that align with an implicit energy function. We also provide model variants that explicitly learn this energy function.

By learning a single equilibrium dynamics, Equilibrium Matching supports \textit{optimization-based sampling}, where samples are obtained by gradient descent on the learned landscape. Unlike existing diffusion samplers that integrate along a prescribed trajectory, optimization-based sampling supports different step sizes and adaptive optimizers. Existing gradient optimization techniques such as Nesterov Accelerated Gradient can be naturally adopted to achieve better generation quality. Equilibrium Matching can also allocate inference-time compute adaptively. It adjusts the sampling steps for each sample independently based on gradient norm and can save up to 60\% of function evaluations while retaining similar generation quality.

We validate Equilibrium Matching through both theoretical analysis and empirical evidence. {Under standard assumptions, we establish that training samples are local attraction points, and that gradient descent converges under a conservative field.} Empirically, EqM achieves {the best FID among the tested methods,} 1.90 on ImageNet 256$\times$256 generation. EqM also exhibits strong scaling behavior, exceeding the flow-based counterpart at all tested scales. These results suggest that EqM is a promising approach for generative modeling.

Beyond generation, EqM demonstrates additional properties {that are not naturally supported by standard diffusion/flow formulations}. EqM can generate high-quality samples directly from partially noised inputs, whereas flow-based models {without the correct noise level} only perform well when starting with pure noise. Moreover, EqM can perform out-of-distribution (OOD) detection without relying on any external module. We also show that different EqM models can be added together to generate compositional images in a similar way as EBMs. {These experiments provide evidence for useful equilibrium-style capabilities.} We hope that Equilibrium Matching provides a principled way to unify flow-based and energy-based generative models and can inspire better inference-time optimizers for generation in the future.

We summarize our contributions as follows:
\begin{itemize}[leftmargin=1.2em,topsep=2pt,itemsep=1pt]
    \item We propose Equilibrium Matching (EqM), a generative framework that learns a time-invariant gradient over an implicit energy landscape.
    \item We introduce optimization-based sampling, where generation is performed by gradient descent rather than integration, offering more flexibility, better performance, and higher efficiency.
    \item {We analyze EqM under idealized population-optimal assumptions, establishing training point attraction and energy-descent guarantees on conservative fields.}
    \item {We empirically validate EqM on image generation, achieving the best FID among the tested methods and competitive on other metrics.}
    \item We show that EqM unlocks capabilities not naturally supported by diffusion/flow models.
\end{itemize}

\begin{figure}[t]
\centering
    \includegraphics[width=0.25\textwidth, trim=2cm 2cm 2cm 2cm, clip]{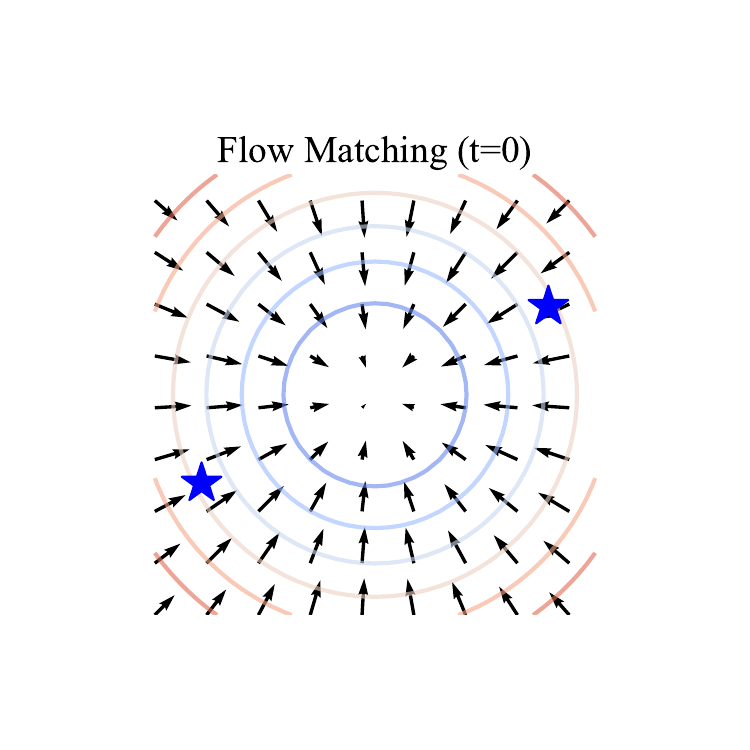}\hhh
    \includegraphics[width=0.25\textwidth, trim=2cm 2cm 2cm 2cm, clip]{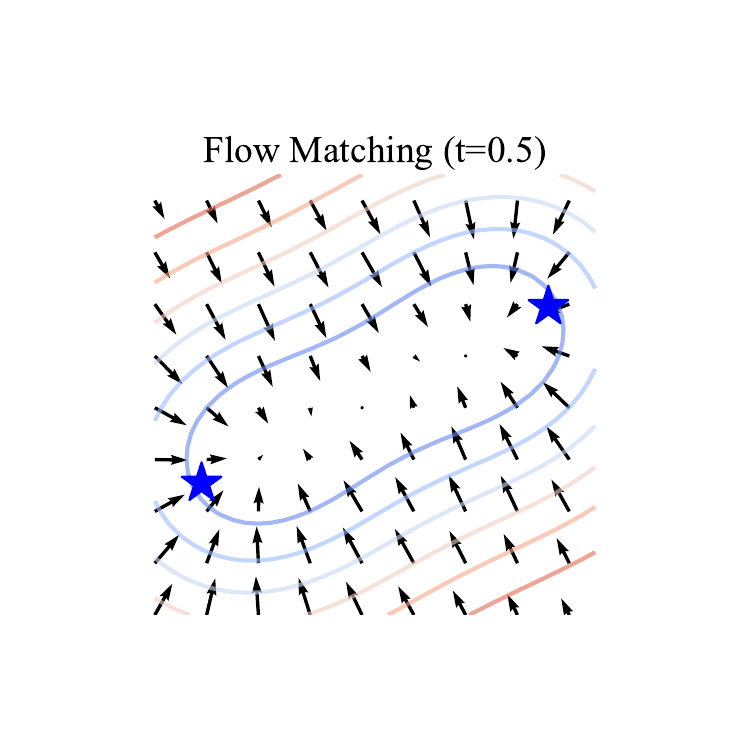}\hhh
    \includegraphics[width=0.25\textwidth, trim=2cm 2cm 2cm 2cm, clip]{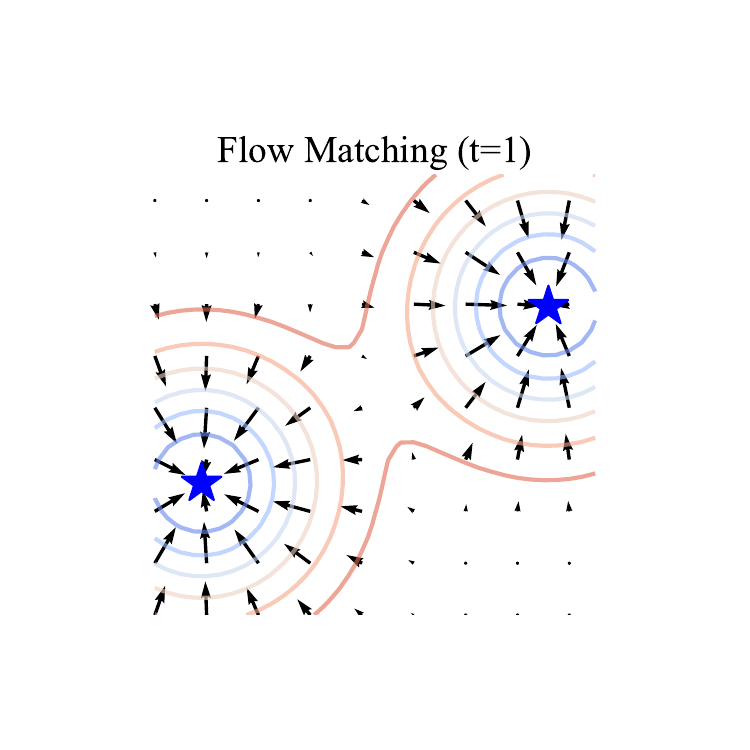}
  \vrule\ 
    \includegraphics[width=0.25\textwidth, trim=2cm 2cm 2cm 2cm, clip]{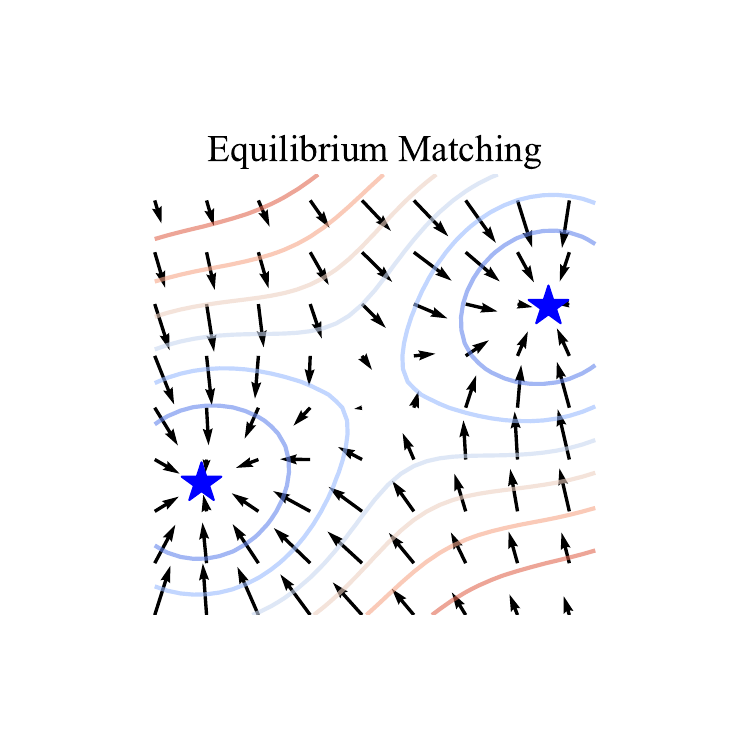}
  \vspace{-5pt}
  \caption{\textbf{Conceptual 2D Visualization.} We compare the conceptual 2D dynamics of Equilibrium Matching and Flow Matching under 2 ground truths (marked by stars). \textbf{Left}: Flow Matching learns \textit{non-equilibrium} velocity that \textit{varies} over time. \textbf{Right}: Equilibrium Matching learns an \textit{equilibrium} gradient that is time-invariant.}
  \label{concept}
  \vspace{-10pt}
\end{figure}

\section{Preliminaries: Flow Matching}
Diffusion and Flow Matching models learn non-equilibrium dynamics. Flow Matching (FM), for example, learns to match the conditional velocity along a linear path connecting noise and image samples. During sampling, Flow Matching starts from pure Gaussian noise and iteratively denoises the current sample using the velocity predicted by $f$. This process is governed by a differential equation framework, in which the predicted velocity is treated as the time derivative of the desired sampling path and integrated over a total length of $1$. Formally, let $f$ denote the generative model, let $\epsilon$ be Gaussian noise, let $x$ be a real image sample from the training dataset, and let $t$ be a timestep sampled uniformly between $0$ and $1$. 
The training objective of FM can be written as:
\begin{equation}
L_\text{FM} = \bigl(f(x_t, t) - (x - \epsilon)\bigr)^2.
\end{equation}
Here $x_t=tx+(1-t)\epsilon$ is the interpolated sample and $x-\epsilon$ is the target velocity. The model $f$ takes both $x_t$ and $t$ as inputs, where $t$ is the corresponding noise level for $x_t$. 

\textbf{Noise-Unconditional Model.} One approach to make Flow Matching learn equilibrium dynamics is to directly remove $t$ as a conditioning input for $f$ \citep{sun2025noise}. In this noise-unconditional version, the model is no longer conditioned on the timestep or noise level, giving a new objective: 
\begin{equation}
L_\text{uncond-FM} = \bigl(f(x_t) - (x - \epsilon)\bigr)^2.
\end{equation}
The only difference is the absence of conditioning on $t$ in $f$. The sampling and training procedures are otherwise identical to those of the original Flow Matching. However, removing noise conditioning like the above degrades generation quality, and this unconditional variant does not exhibit additional properties that differ from those of the original model.

\section{Equilibrium Matching}
  
Equilibrium Matching (EqM) learns a time-invariant gradient field that is {designed to be compatible with} an underlying energy function, eliminating time/noise conditioning and fixed-horizon integrators. Conceptually (\cref{concept}), EqM's gradient vanishes on the data manifold and increases toward noise, yielding an equilibrium landscape in which ground-truth samples are stationary points. Flow Matching learns a varying velocity that only converges to ground truths at the final timestep, whereas EqM attempts to learn a time-invariant gradient landscape. 

\subsection{Training}
  
To train an Equilibrium Matching model, we aim to construct an energy landscape in which the target gradient at ground-truth samples is zero. To do so, we first define a corruption scheme that provides a transition between data and noise. Let $\gamma$ be an interpolation factor sampled uniformly between $0$ and $1$, let $\epsilon$ be Gaussian noise, and let $x$ be a sample from the training set. Denote $x_\gamma=\gamma x+(1-\gamma)\epsilon$ as an intermediate corrupted sample. Unlike $t$ in FM, our $\gamma$ is implicit and not seen by the model. Our goal is to define a target gradient at these intermediate samples $x_\gamma$ that matches an implicit energy landscape. Using a gradient direction that descends from noise to data, we write the Equilibrium Matching training objective as:
\begin{equation}
L_\text{EqM}=\bigl(f(x_\gamma)-(\epsilon-x)c(\gamma)\bigr)^2,
\end{equation}
where $c(\gamma)$ controls the gradient magnitude. The target gradient $(\epsilon-x)c(\gamma)$ has direction $(\epsilon-x)$ and magnitude $c(\gamma)$ that vanishes as we get closer to the data manifold. We explicitly enforce $c(1)=0$, ensuring that the energy landscape has vanishing gradients at real samples. Together, the direction and magnitude result in a target gradient that supports an implicit energy landscape. When $c(\gamma)=1$, the EqM target is exactly the negation of FM's target.

Compared with Flow Matching, EqM's objective is derived from an EBM perspective rather than a normalizing flow's perspective. More specifically, the EBM interpretation suggests that generation is an optimization procedure where we can use an arbitrary amount of compute to find the minima, as opposed to flow matching, where we have to integrate along a fixed trajectory from the noise to the sample.  This difference in perspectives results in different fundamental constraints. EqM requires $c(1)=0$ to construct an energy landscape with local minima at the data manifold, whereas FM requires $\int_{\gamma=0}^1 c(\gamma)=1$ to construct a valid integration path. Our interpretation also results in a different direction of target, where EqM learns the gradient $\epsilon-x$ and FM learns the velocity $x-\epsilon$.  We next discuss several simple choices for $c$ in EqM.

\textbf{Linear Decay.} A natural choice for $c$ is a linear function that decays from $1$ to $0$. In the energy landscape, this is equivalent to assigning noise a high gradient and making the gradient decay linearly to $0$ toward the ground-truth image:
\begin{equation}
    c_\text{linear}(\gamma) = 1-\gamma.
\end{equation}

\textbf{Truncated Decay.} Beyond linear decay, we may want the gradient to remain constant when far away from data. This leads to a truncated decay, where the target gradient stays at $1$ when $\gamma \leq a$ ($a\in[0,1)$) and then decays linearly to 0 when approaching the data manifold:
\begin{equation}
    c_\text{trunc}(\gamma) = \begin{cases}
  1, & \gamma \leq a\\
   \frac{1-\gamma}{1-a}, & \gamma > a \end{cases}.
\end{equation}

\textbf{Piecewise.} We can also vary the constant segment of the truncated decay function and set its starting value to $b$, with $b\in[0,\infty)$. This gives a piecewise function that starts at $b$, changes linearly to 1 at $\gamma=a$, and then decays linearly down to 0 at $\gamma=1$:
\begin{equation}
        c_\text{piece}(\gamma) = \begin{cases}
  b-\frac{b-1}{a}\gamma, & \gamma \leq a\\
   \frac{1-\gamma}{1-a}, & \gamma > a \end{cases}.
\end{equation}

\textbf{Gradient Multiplier.} The above choices for $c$ have varying magnitudes. Thus, we introduce an additional gradient multiplier $\lambda$ on top of these gradient fields to control the overall scale. Using linear decay as an example, the final function becomes $c(\gamma) = \lambda c_\text{linear}(\gamma) = \lambda(1-\gamma)$.

\subsection{Learning Explicit Energy}
Previously, we treated the energy landscape as an implicit underlying structure and learned the gradient of this implicit energy function. We can also modify the Equilibrium Matching model to learn explicit energy values, which we use primarily as a diagnostic. For an explicit energy model $g$, we match the gradient $\nabla g(x_\gamma)$ at corrupted samples and take $E=g(x_\gamma)$ as the energy at $x_\gamma$. This naturally constructs an energy landscape in which real samples are assigned low energy while noisy samples are assigned high energy. The Equilibrium Matching with Explicit Energy (EqM-E) objective can be written as:
\begin{equation}
    L_\text{EqM-E}=(\nabla g(x_\gamma) - (\epsilon-x)c(\gamma))^2.
\end{equation}
$\nabla g(x_\gamma)$ is the derivative of $g$ with respect to input $x_\gamma$. To obtain a scalar function $g$, we follow \cite{du2023reduce} and discuss two different ways to construct $g$ from an existing Equilibrium Matching model $f$ without having to introduce new parameters. 

\textbf{Dot Product.} The first approach uses the dot product between the input $x_\gamma$ and the output $f(x_\gamma)$, defined as $g(x_\gamma) = x_\gamma \cdot f(x_\gamma)$. The corresponding derivative with respect to $x_\gamma$ is {$\nabla g(x_\gamma) = f(x_\gamma) + J_f(x_\gamma)^T x_\gamma$}, where $J_f\in\mathbb R^{d\times d}$ is the Jacobian of $f$.

\textbf{Squared $L_2$ Norm.} The second approach uses the squared $L_2$ norm of the output $f(x_\gamma)$ with a factor of one half to simplify coefficients: $g(x_\gamma) = -\frac{1}{2}||f(x_\gamma)||_2^2$. The corresponding derivative is {$\nabla g(x_\gamma) = -J_f(x_\gamma)^T f(x_\gamma)$}.

\subsection{Sampling}
Because of its equilibrium nature, EqM generates samples via optimization on the learned landscape. In contrast to diffusion/flow models that integrate over a fixed time horizon, EqM decouples sample quality from a prescribed trajectory. It formulates sampling as a gradient descent procedure and supports adaptive step sizes, optimizers, and compute.

\textbf{Gradient Descent Sampling (GD).} \label{stepsize} A simple way to sample from an EqM model is to apply vanilla gradient descent. Let $x_k$ denote the sample after $k$ steps, then an update step with step size $\eta$ is:
\begin{equation}
    x_{k+1} \leftarrow x_k - \eta\nabla E(x_k),
\end{equation}
where $\nabla E(x_k)$ is the predicted gradient at $x_k$ and $E$ may be learned implicitly ($\nabla E(x)=f(x)$) or explicitly ($\nabla E(x) = \nabla g(x)$).

\textbf{Sampling with Nesterov Accelerated Gradient (NAG-GD).} Building on gradient descent sampling, we can adopt existing optimization techniques in our sampling procedure. As an example, we use Nesterov Accelerated Gradient \citep{nesterov1983method}, which applies a look-ahead step at each update and evaluates the gradient at that look-ahead point:
\begin{equation}
    x_{k+1} \leftarrow x_k - \eta\nabla E(x_k+\mu(x_k-x_{k-1})),
\end{equation}
where $\mu$ is the look-ahead factor controlling how far to look ahead at each step.

\textbf{Sampling with Differential Equations.} EqM also naturally supports integration-based samplers. ODE-based diffusion samplers can be viewed as a special case of our gradient-based method. We discuss this further in \cref{odeeqv}.

\textbf{Sampling with Adaptive Compute.} Another advantage of gradient-based sampling is that instead of a fixed number of sampling steps, we can allocate adaptive compute per sample by stopping when the gradient norm drops below a certain threshold $g_\text{min}$. For step size $\eta$ and threshold $g_\text{min}$, we perform sampling steps $x_{k+1} \leftarrow x_k - \eta\nabla E(x_k)$ until $||\nabla E(x_k)||_2 < g_\text{min}$. This mechanism allows the sampler to adaptively adjust the number of steps for each sample and terminate automatically when the current sample is already close enough to the data manifold.

\subsection{Implementation}
Equilibrium Matching is simple to implement. We provide example pseudocode for training in \cref{code} and sampling in \cref{code2}. During training, we first compute an interpolated corrupted sample $x_\gamma$ from noise $\epsilon$, image $x$, and factor $\gamma$, then train the model $f$ to predict the target gradient $(\epsilon - x)c(\gamma)$ by minimizing a mean squared error objective. At inference time, Equilibrium Matching uses the predicted gradients to iteratively optimize via gradient descent. 

We adopt the transformer-based backbone from \cite{sit} for our EqM model, using the exact implementation to ensure that no architectural differences influence the results. To remove conditioning on $t$ from this backbone, we set the input $t$ to $0$. For further details on model configurations, see \cref{expset}.
\begin{table}[t]
  \vspace{-10pt}
  \centering
   \begin{minipage}[t]{0.48\columnwidth}
  \begin{algorithm}[H]
\caption{EqM Training\\ \scriptsize{The loss function takes as input model $f$, noise $\epsilon$ (eps), image $x$, and an interpolation factor $\gamma$ (g), and returns the EqM loss.}}
\begin{lstlisting}[belowskip=-25pt,aboveskip=0pt]
def training_loss(f, eps, x, g):
  xg = (1-g) * eps + g * x
  target = (eps-x) * c(g)
  loss = (f(xg) - target)**2
  return loss
  \end{lstlisting}
  \label{code}
  \end{algorithm}
\end{minipage}\hfill
\begin{minipage}[t]{0.48\columnwidth}
\begin{algorithm}[H]
\caption{EqM Sampling\\ \scriptsize{The sampling function takes as input pretrained model $f$, initialization $st$, step size $\eta$, and total steps $N$, and returns the generated sample.}}
\begin{lstlisting}[belowskip=-25pt,aboveskip=0pt]
def generate(f, st, eta, N):
  xn = st
  for i in range(N):
    xn = xn - eta*f(xn)
  return xn
  \end{lstlisting}
  \label{code2}
  \end{algorithm}
  \end{minipage}
  \vspace{-15pt}
  \end{table}

\section{Analysis}
{We give an idealized population-optimal analysis of EqM and characterize critical points of the learned field.}

\textbf{Assumptions.} {Let $f^*$ denote the continuous version of the population-optimal EqM field on a finite training set $\mathcal X\subset\mathbb R^d$ with $d\ge3$. We assume $c(1)=0$ and $c(\gamma)\le C(1-\gamma)$ near $\gamma=1$. For local-stability statements we specialize to linear decay $c(\gamma)=\lambda(1-\gamma)$. An exact energy-descent result additionally assumes a conservative field $f=\nabla E$, which holds by construction for EqM-E; for implicit EqM we only verify approximate conservativity empirically.}

Under these assumptions, we can derive the following statements (see Appendix for derivations and scope):

\begin{statement}[Training samples are critical points]\label{st1}
{For every $x^{(i)}\in\mathcal X$, the population-optimal field satisfies $f^*(x^{(i)})=0$.} (Derivation in \cref{st1der})
\end{statement}

\begin{statement}[Local attraction and scope of non-data equilibria]\label{st11}
{For linear decay, training samples are locally attracting critical points and every zero of $f^*$ lies in $\operatorname{conv}(\mathcal X)$. A non-data zero is provably unstable under the explicit $(k,\kappa)$-resolvability assumption of \cref{st11der}. (Derivation in \cref{st11der})}
\end{statement}

\begin{statement}[Gradient descent on an explicit energy converges to stationarity]\label{st2}
{For an explicit-energy EqM-E field $f=\nabla E$, if $E$ is $L$-smooth and bounded below, gradient descent with $\eta\in(0,1/L]$ satisfies}
\[
\min_{0\leq k<K}\|f(x_k)\|^2 \leq \frac{2\bigl(E(x_0)-E_\text{inf}\bigr)}{\eta K}.
\]
(Derivation in \cref{st2der})
\end{statement}

{Statements~\ref{st1} and~\ref{st11} analyze the population-optimal field, while Statement~\ref{st2} relies on a conservative or explicit-energy field. For the trained implicit EqM network, we verify empirically later that the gradient field is roughly conservative in practice.}

\textbf{Empirical Interpretation.} We emphasize that these results \textit{do not} imply EqM only memorizes training samples in practice. Similar theoretical conclusions hold for diffusion and Flow Matching, whose generalizations are established empirically. We likewise demonstrate generalization for EqM empirically in \cref{nn} of \cref{imagen}. {Complementary precision/recall, density/coverage, and DINOv2-feature metrics are given in \cref{rebuttalcontrols}.} The implicit-field association with an energy is {approximately true in practice}, as quantified in \cref{tab:conserv}.

\section{Experiments}
We validate the practical performance of EqM from four major perspectives. First, we demonstrate the advantages in image generation quality and generalization in practice. Then, we examine the properties and performance of our gradient-based sampling methods, compared against traditional integration-based methods. Next, we illustrate the effectiveness and validity of our gradient landscape design via ablation studies. Finally, we show additional properties of EqM that are not inherently supported by diffusion/flow methods. For details of the training and sampling settings, see \cref{expset}.
\begin{figure}[t]
\begin{minipage}[b]{0.3\textwidth}
  \centering    
\includegraphics[width=0.33\linewidth]{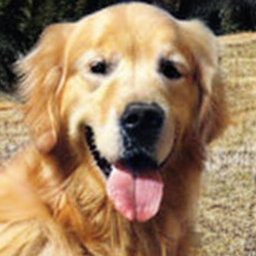}\hh
\includegraphics[width=0.33\linewidth]{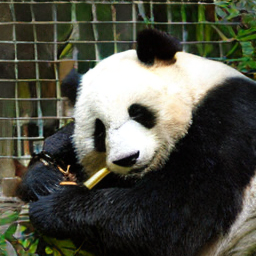}\hh
\includegraphics[width=0.33\linewidth]{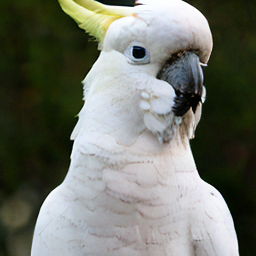}\\ \vspace{-0.1em}
\includegraphics[width=0.33\linewidth]{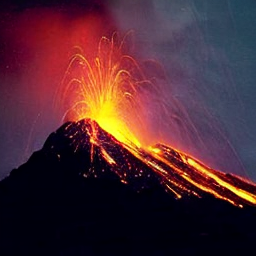}\hh
\includegraphics[width=0.33\linewidth]{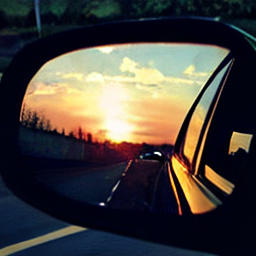}\hh
\includegraphics[width=0.33\linewidth]{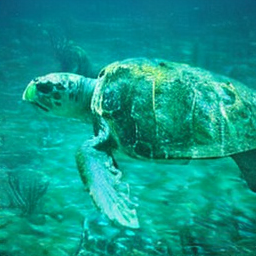}\\
\vspace{-5pt}
  \captionof{figure}{\textbf{Curated Samples.} We present curated samples generated by our EqM-XL/2 model.}
  \label{visuals}  
\end{minipage}\hfill
\begin{minipage}[b]{0.68\textwidth}
  \centering
      \includegraphics[scale=0.235]{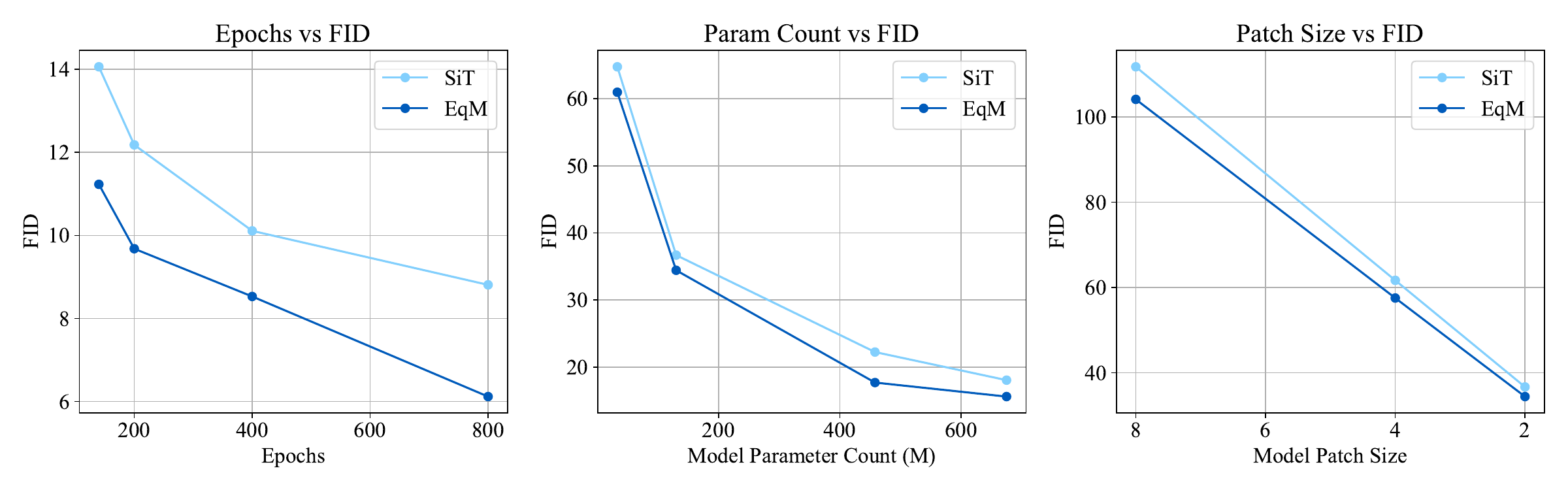}
\vspace{-15pt}
  \captionof{figure}{\textbf{Scalability of Equilibrium Matching.} EqM scales across training epochs (left), parameter count (middle), and patch size (right), and outperforms Flow Matching at all tested scales by a significant margin. }
    \label{scale}
\end{minipage}
\vspace{-5pt}
\end{figure}
\begin{figure}[t]
\begin{minipage}[b]{0.42\textwidth}
  \centering
  \small
\begin{tabular}{lcccc}
\toprule
model & method &  FID \\
\midrule
StyleGAN-XL &GAN& 2.30\\
VDM++ & Diffusion  & 2.12 \\
DiT-XL/2 & Diffusion  & 2.27\\
SiT-XL/2 &  FM  & 2.06\\
\textbf{EqM-XL/2} & \textbf{EqM}  & \textbf{1.90} \\
\bottomrule
\end{tabular}
\vspace{-5pt}
\captionof{table}{\textbf{Class-Conditional ImageNet 256$\times$256 Generation.} {EqM-XL/2 achieves the best FID among the tested methods (1.90).}}
\label{main}
\end{minipage}\hfill
\begin{minipage}[b]{0.54\textwidth}
  \centering
  \small
\begin{tabular}{lcccc}
\toprule
model & sampler & $\eta$ & $\mu$ & FID\\
\midrule
SiT-XL/2 & Euler {\scriptsize (ODE)}  & 0.0040 & - & 2.10 \\
SiT-XL/2 & Heun {\scriptsize (SDE)}  & 0.0040 & - &2.06 \\
EqM-XL/2 & Euler {\scriptsize (ODE)}  & 0.0017 & - &1.93 \\
EqM-XL/2 & GD  & 0.0017 & - &1.93 \\
\textbf{EqM-XL/2} & NAG-GD & 0.0017 & 0.3 &\textbf{1.90} \\
\bottomrule
\end{tabular}
\vspace{-5pt}
\captionof{table}{\textbf{Sampler Comparison.} {EqM achieves lower FID than the tested Flow Matching baseline} using both integration-based ODE sampler and gradient-based samplers. }
\label{samplerabl}
\end{minipage}\hfill
\vspace{-10pt}
\end{figure}
\subsection{Image Generation}
\label{imagen}
  
\textbf{ImageNet Results.} We report performance on class-conditional ImageNet \citep{imgnet} 256$\times$256 image generation. We compare Equilibrium Matching with prior generative methods, including StyleGAN \citep{sauer2022stylegan}, VDM++ \citep{kingma2023understanding}, DiT \citep{dit}, and SiT \citep{sit}. Results are shown in \cref{main}. EqM achieves an FID of $1.90$, {the best FID among the tested methods}. {On additional metrics, EqM-XL/2 obtains sFID/IS $4.54/275.7$ versus $4.49/277.5$ for SiT-XL/2. Additional precision/recall, density/coverage, and DINOv2-feature results are reported in \cref{rebuttalcontrols}.} We also plot the evolution of FID over time while training our EqM-XL/2 model and compare it with the training FID curve of SiT-XL/2 in \cref{scale}. EqM consistently improves over the Flow Matching baseline throughout training, {showing a consistent FID advantage in the tested ImageNet settings}.

\begin{figure}[t]
\begin{minipage}[t]{0.48\textwidth}
\vspace{0em}
  \centering    
  \begin{minipage}{0.1\linewidth}
  \centering
    FM
  \end{minipage}
  \includegraphics[width=0.13\linewidth]{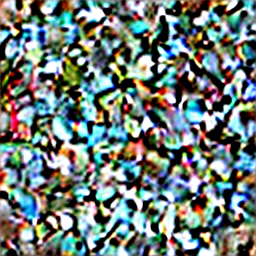}\hh
\includegraphics[width=0.13\linewidth]{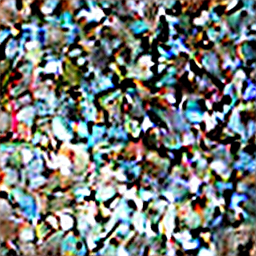}\hh
\includegraphics[width=0.13\linewidth]{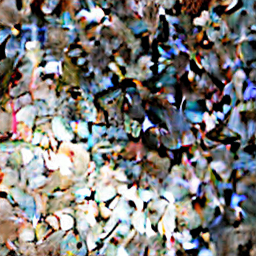}\hh
\includegraphics[width=0.13\linewidth]{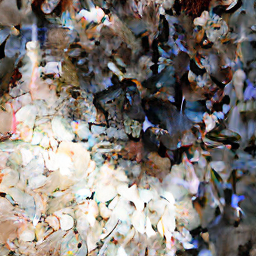}\hh
\includegraphics[width=0.13\linewidth]{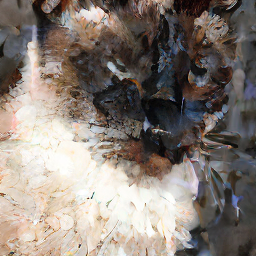}\hh
\includegraphics[width=0.13\linewidth]{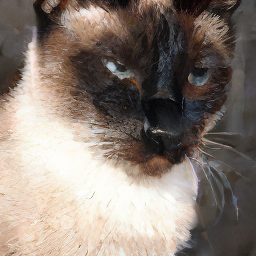}\hh
\includegraphics[width=0.13\linewidth]{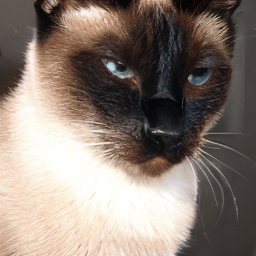}\hh\\
\begin{minipage}{0.1\linewidth}
  \vspace{-1.5em}
  \centering
    EqM
  \end{minipage}
\includegraphics[width=0.13\linewidth]{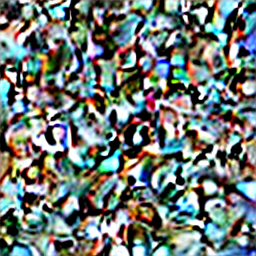}\hh
\includegraphics[width=0.13\linewidth]{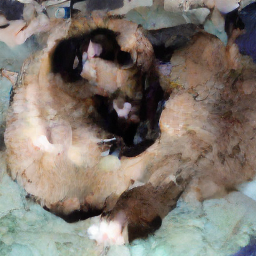}\hh
\includegraphics[width=0.13\linewidth]{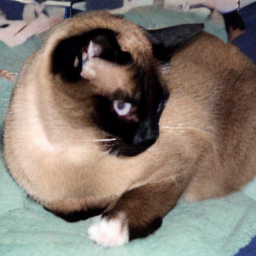}\hh
\includegraphics[width=0.13\linewidth]{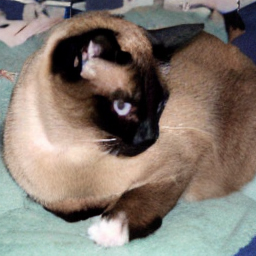}\hh
\includegraphics[width=0.13\linewidth]{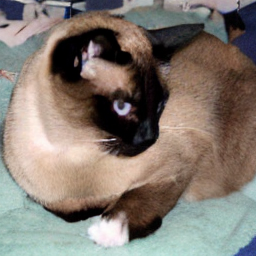}\hh
\includegraphics[width=0.13\linewidth]{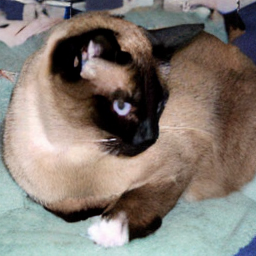}\hh
\includegraphics[width=0.13\linewidth]{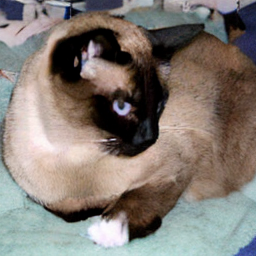}\\
\vspace{-5pt}
  \captionof{figure}{\textbf{Sampling Process Visualization.} We present intermediate samples from XL/2 models using matched average step displacement. EqM (bottom) produces realistic images earlier than FM (top).}
  \label{sampvis}
\end{minipage}\hfill
\begin{minipage}[t]{0.48\textwidth}
\vspace{-0.5em}
  \centering    
\begin{minipage}[t]{0.48\textwidth}
\vspace{0em}
\centering
\hspace{-2.7em}samples\hspace{2.0em}NNs
\end{minipage}\hfill
\begin{minipage}[t]{0.48\textwidth}
\vspace{0em}
\centering
\hspace{-3.2em}samples\hspace{2.0em}NNs
\end{minipage}\\
\includegraphics[width=0.12\linewidth]{dog.png}\hspace{0.1em}
\includegraphics[width=0.12\linewidth]{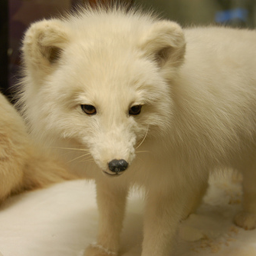}\hh
\includegraphics[width=0.12\linewidth]{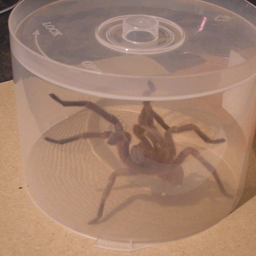}\hh
\includegraphics[width=0.12\linewidth]{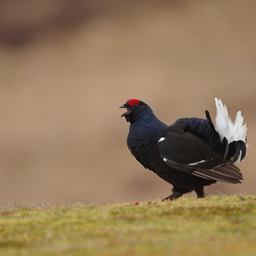}\hspace{0.3em}
\includegraphics[width=0.12\linewidth]{bird.png}\hspace{0.1em}
\includegraphics[width=0.12\linewidth]{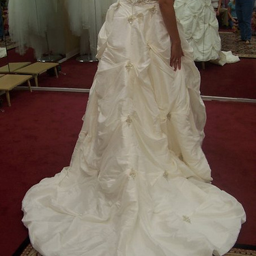}\hh
\includegraphics[width=0.12\linewidth]{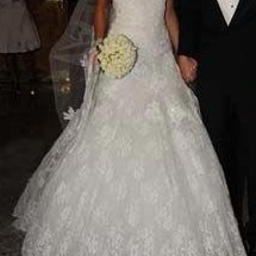}\hh
\includegraphics[width=0.12\linewidth]{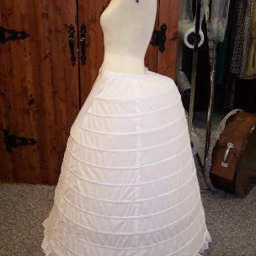}\\

\includegraphics[width=0.12\linewidth]{volcano.png}\hspace{0.1em}
\includegraphics[width=0.12\linewidth]{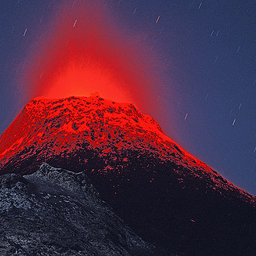}\hh
\includegraphics[width=0.12\linewidth]{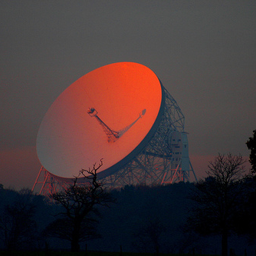}\hh
\includegraphics[width=0.12\linewidth]{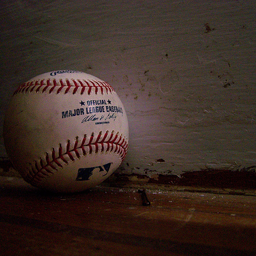}\hspace{0.3em}
\includegraphics[width=0.12\linewidth]{turtle.png}\hspace{0.1em}
\includegraphics[width=0.12\linewidth]{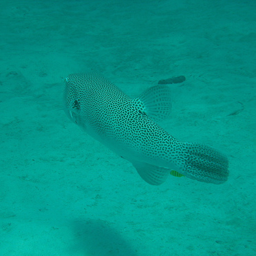}\hh
\includegraphics[width=0.12\linewidth]{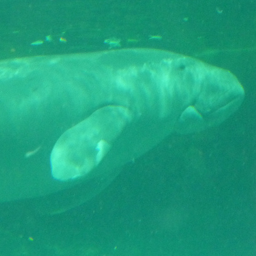}\hh
\includegraphics[width=0.12\linewidth]{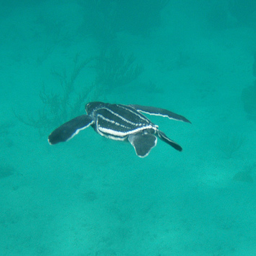}\\
\vspace{-5pt}
\captionof{figure}{\textbf{Nearest Neighbors (NNs) of EqM Samples in the Training Set.} EqM produces samples that are not in the training set, suggesting that it generalizes and does not only memorize the training images.}
  \label{nn}
\end{minipage}
\vspace{-10pt}
\end{figure}

\textbf{Visualizations.} We present curated samples from our EqM-XL/2 model in \cref{visuals} and visualizations of the sampling process in \cref{sampvis}. For both EqM and FM, we use XL/2 models trained for 1400 epochs to produce the visualizations. In \cref{sampvis}, we observe that EqM converges faster than its FM counterpart at inference. {To control for effective update scale, we match the default mean per-step displacement for EqM and FM, so the qualitative difference is not explained only by update magnitude.} In \cref{nn}, we show the top-3 nearest neighbors (measured by mean squared distance) in the training set for EqM-generated samples. The nearest neighbors differ from the generated samples, indicating that EqM does not memorize or overfit to the training dataset and can generalize to produce previously unseen samples at inference time.

\textbf{Scalability.} To assess scalability, we vary training length, model size, and patch size. We report the evolution of FID over training using the XL/2 training curves. For model-size scaling, we fix the patch size to $2$ and the number of training epochs to $80$, and evaluate all four model sizes: S, B, L, and XL. For patch-size scaling, we fix the model size to B and training to $80$ epochs, and evaluate patch sizes $8$, $4$, and $2$. As shown in \cref{scale}, EqM scales well along all axes and consistently outperforms Flow Matching under all tested configurations. These results suggest that EqM has strong scaling potential and is a promising alternative to Flow Matching.

\begin{figure}[t]
\begin{minipage}[t]{0.31\textwidth}
\centering
    \includegraphics[scale=0.27]{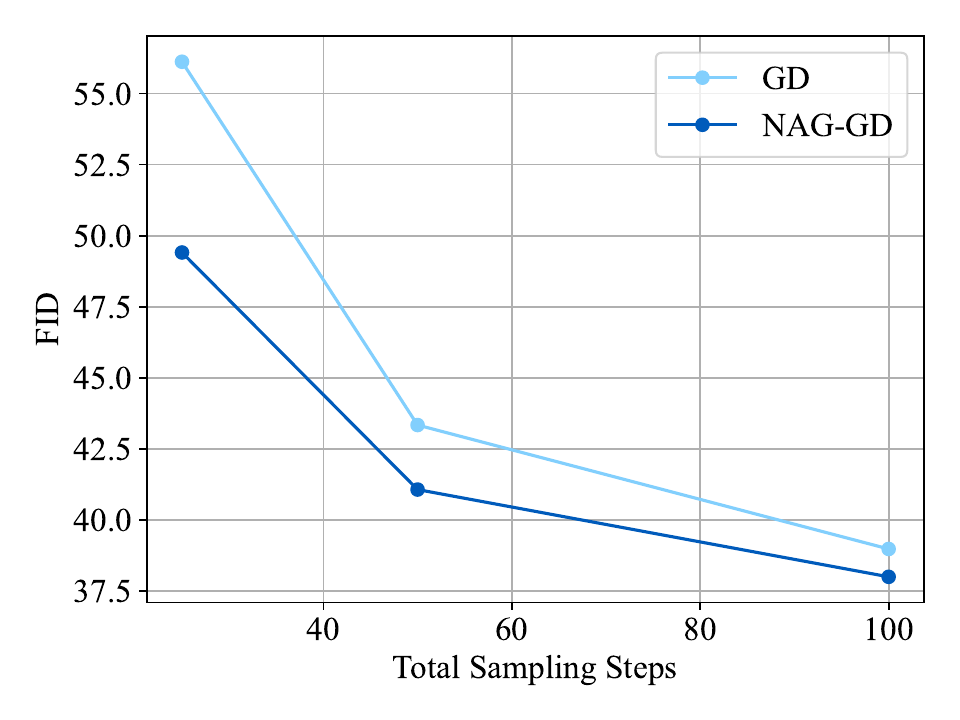}
    \vspace{-18pt}
  \captionof{figure}{\textbf{Sampling with Nesterov Accelerated Gradient.} NAG-GD achieves better sample quality than GD, with the gap being more significant when using fewer steps.}
  \label{nag}
\end{minipage}
\hfill
\begin{minipage}[t]{0.31\textwidth}
\centering
    \includegraphics[scale=0.27]{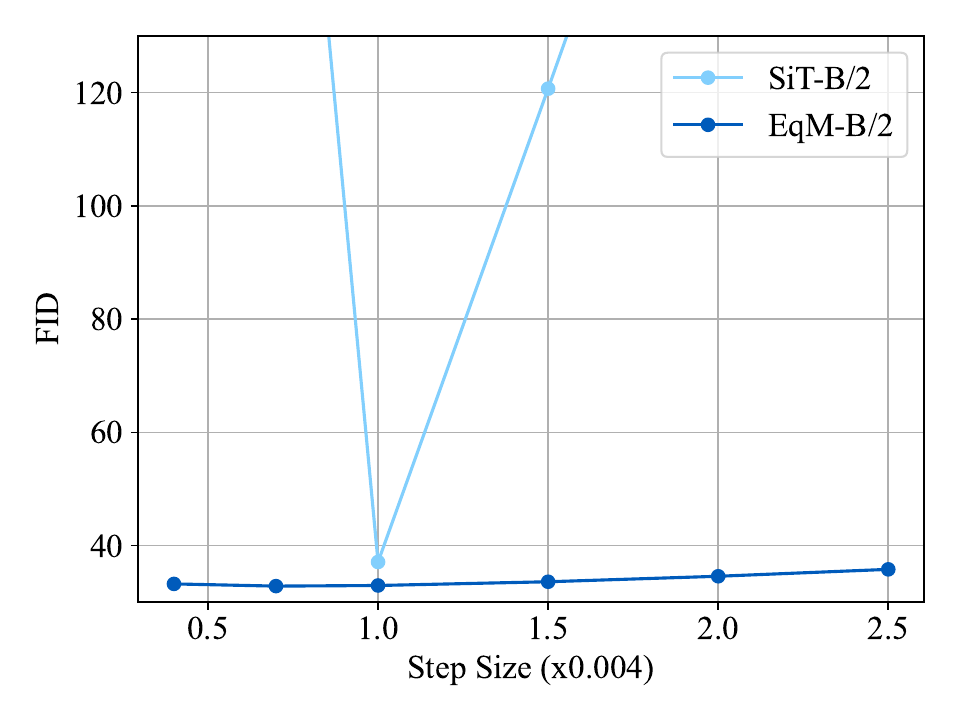}
    \vspace{-18pt}
  \captionof{figure}{\textbf{Different Sampling Step Sizes.} EqM is robust to a wide range of step sizes, whereas Flow Matching only functions properly at one specific step size.}
    \label{step}
\end{minipage}
\hfill
\begin{minipage}[t]{0.31\textwidth}
  \centering
\includegraphics[scale=0.23]{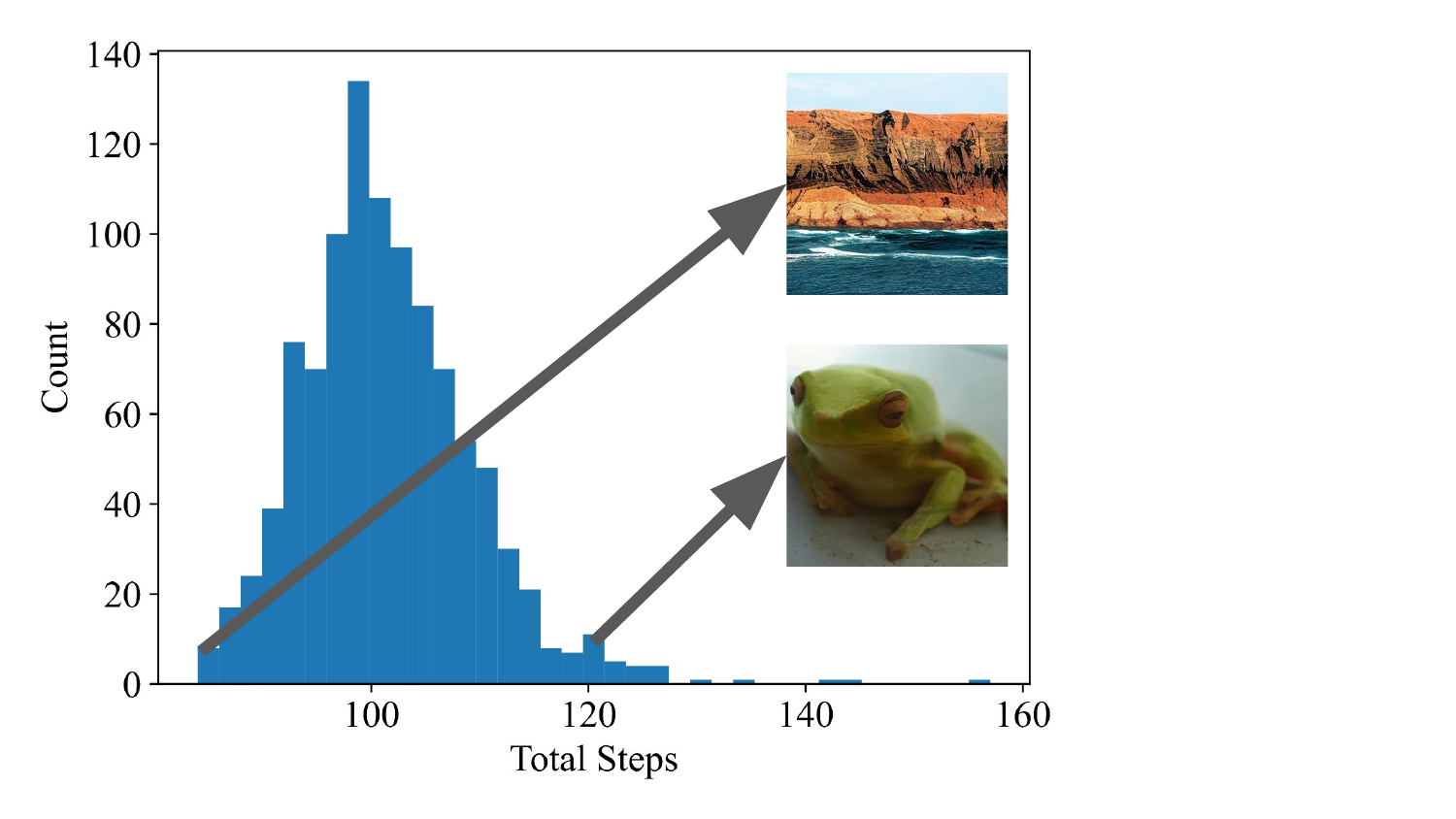}
    \vspace{-18pt}
\captionof{figure}{\textbf{Total Steps Under Adaptive Compute.} EqM assigns different numbers of steps for different samples, adaptively adjusting compute at inference time. }
\label{adap}
\end{minipage}
\vspace{-15pt}
\end{figure}

\subsection{Inference-Time Experiments}
\label{opt}
  
\textbf{Different Gradient Samplers.} We evaluate our proposed gradient-based samplers on ImageNet generation using the EqM-B/2 model. For NAG-GD, we use $\mu=0.35$, which we found empirically to work best. As shown in \cref{nag}, NAG-GD yields improved FID across all tested numbers of steps. Moreover, the quality gap increases as the total number of steps decreases. This aligns with our intuition that with fewer total steps, gradient descent requires more assistance to reach a desirable local minimum, making NAG more effective. In \cref{samplerabl}, we also compare traditional integration samplers with the proposed gradient descent samplers on EqM-XL/2. Euler ODE sampler, plain gradient descent, and NAG-GD all exceed the Flow Matching baseline by a large margin, with NAG-GD performing the best. These results show that the NAG technique, commonly used in optimization, is also effective during sampling in Equilibrium Matching, providing further evidence that our approach enables new opportunities at inference time. We also investigate adaptive and stochastic optimizers in \cref{moreoptzer}.

\textbf{Flexible Step Size.} Viewing sampling through an optimization perspective implies that the step size can be adjusted freely. We evaluate this claim by varying the sampling step size on EqM-B/2. For comparison, we also report the performance of the FM baseline, where we use $\eta$ to replace the ODE update length at each step. We use a total of $N=250$ sampling steps. From \cref{step}, we observe that our EqM model’s generation quality remains high and exceeds the Flow Matching baseline across all tested step sizes. By contrast, Flow Matching requires a specific step size of $\eta=0.004=\frac{1}{N}$ to function properly, and small fluctuations in step size lead to significantly worse performance. This suggests that EqM constructs a fundamentally different landscape than FM, which enables new sampling schemes not supported by FM models.

\textbf{Adaptive Compute.} We test our adaptive compute sampling using a gradient norm threshold of 10 on the EqM-B/2 model. We observe that EqM is able to generate reasonably good samples by adaptive compute, achieving a reasonable FID of 33.79 (32.85 without adaptive compute) using the EqM-B/2 model. We present the distribution of total sampling steps for 1024 samples in \cref{adap}, which suggests that EqM assigns different inference-time compute for different samples and manages to lower the total compute to 40\% of the original compute (original sampling uses fixed 250 steps). Our results offer promising evidence that EqM can enable new inference-time improvements.

\subsection{Ablation Study}
\label{ablation}
  
\begin{figure}
\begin{minipage}[b]{0.65\linewidth}
\begin{center}
\small
\begin{tabular}{cccccccc}
\toprule
$c(\gamma)$ & \makecell{constant\\(uncond FM \citep{sun2025noise})} & linear & \multicolumn{3}{c}{truncated} & \multicolumn{2}{c}{piecewise} \\
\midrule
$a$ & - & - &0.5 & 0.8 & 0.9 & 0.8 & 0.8\\
$b$ & - & - &- & - & - & 0.8 & 1.4\\
\midrule
FID & 40.81& 50.47 & 38.98 & \textbf{38.34} & 41.22& 38.84& 38.75 \\
\bottomrule
\end{tabular}
\end{center}
\vspace{-8pt}
\captionof{table}{\textbf{Different Training Targets.} Several settings exceed the noise unconditional Flow Matching baseline in performance. Best performance achieved with the truncated decay $c_\text{trunc}$ and hyperparameter $a=0.8$.}
\label{traj}
\end{minipage}\hfill
\begin{minipage}[b]{0.33\linewidth}
\centering
    \includegraphics[scale=0.25]{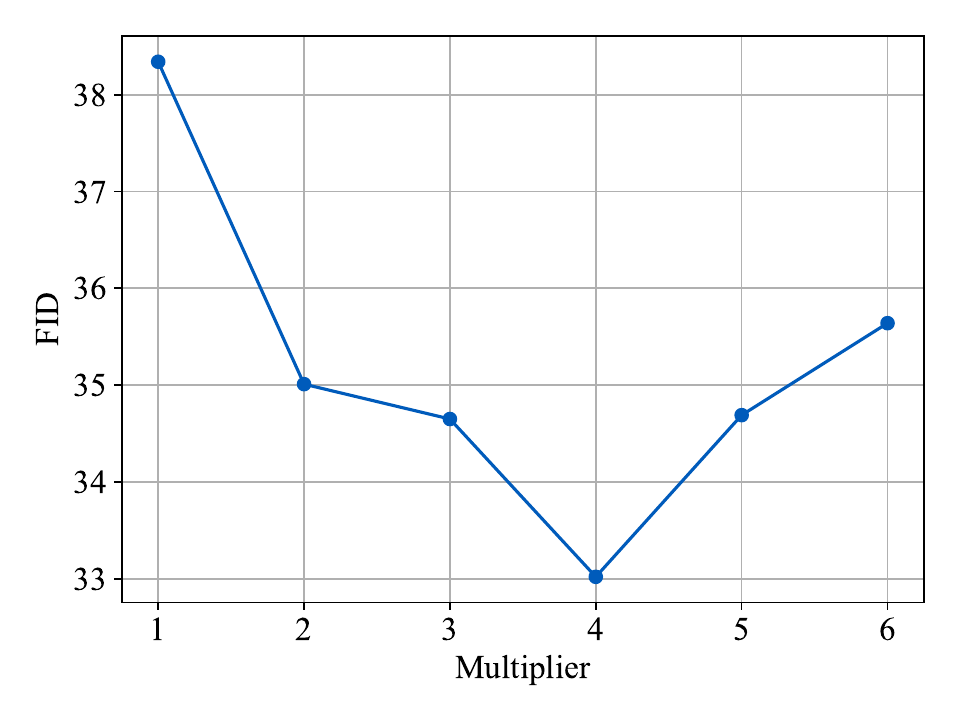}
\vspace{-10pt}
     \captionof{figure}{\textbf{Different Gradient Multipliers.} $\lambda=4$ performs the best.}
    \label{mult}
\end{minipage}
\vspace{-5pt}
\end{figure}
\begin{figure}
\begin{minipage}[b]{0.47\textwidth}
\begin{center}
\small
\begin{tabular}{ccccc}
\toprule
 & FM & \makecell{FM w/ \\EqM target}  & \makecell{uncond\\FM  \citep{sun2025noise}}  & EqM \\
\midrule
\makecell{t-cond} & yes & yes & no & no\\
$c(\gamma)$ & const & trunc & const & trunc \\
\midrule
FID & 36.68 & 41.89 & 40.81 & \textbf{32.85} \\
\bottomrule
\end{tabular}
\vspace{-5pt}
\captionof{table}{\textbf{Noise Conditioning Ablation.} Our truncated decay $c_\text{trunc}$ improves generation only when not conditioned on noise level, matching our hypothesis. }
\label{cond}
\end{center}
\end{minipage}\hfill
\begin{minipage}[b]{0.5\textwidth}
    \centering
    \small

    \begin{tabular}{ccccc}
    \toprule
    model & FM  & FM  & EqM  & EqM \\
        \midrule
    sampler & Euler  & Heun  & GD & NAG-GD  \\
    steps  & 250 & 250 & 250 & 250  \\
        \midrule
    \makecell{wall-clock time}  & 70s & 272s & 59s & 55s \\
    FID & 2.10 & 2.06 & 1.93 & 1.90 \\
    \bottomrule
    \end{tabular}
    \vspace{-5pt}
    \captionof{table}{{  \textbf{Sampling Time Comparison.} We compare the wall-clock time for different sampling methods and find EqM is time-efficient against FM.}}
    \label{samptime}
\end{minipage}
\vspace{-15pt}
\end{figure}
\textbf{Hyperparameter Choices.} To determine the best target landscape, we search over choices and hyperparameters for $c(\gamma)$. We use our EqM-B/2 model and 80 epochs of training for all hyperparameter experiments. We tune the step size of GD sampler and report the best result for each setting in \cref{traj}. The best-performing gradient landscape is truncated decay with $a=0.8$. Our results suggest that it is helpful to keep a constant target gradient at the start of the trajectory before decaying to 0. We then sweep the gradient multiplier $\lambda$ on this best-performing gradient field. As shown in \cref{mult}, a multiplier of $4$ significantly improves FID. We use this setting ($c_{\text{trunc}}$ with $a=0.8$ and $\lambda=4$) as default in our experiments.

\textbf{Noise Conditioning.} We compare our new target gradient with the standard Flow Matching target under both noise-conditional and noise-unconditional settings in \cref{cond}. We note that two specific ablation settings in \cref{cond} correspond exactly to FM and unconditional FM \citep{sun2025noise}. We adopt the 80 epochs B/2 model for both the FM baseline and EqM. Our target gradient improves performance only in the noise-unconditional case, which aligns with our expectation that an energy landscape with zero gradient at real samples is more favorable under equilibrium dynamics.

\textbf{Sampling Time.} {  For a more quantitative analysis on sampling compute, we report the wall-clock time required for each sampling approach in \cref{samptime}. We use the XL/2 model to generate 50,000 samples using identical batch size for all methods, and we report the averaged time per batch of 64 samples on a single H100 GPU. We find that EqM not only achieves better FID but also samples faster than typical differential-equation-based samplers. {At equal NFE, part of this difference comes from EqM's time-independent conditioning path (the timestep embedding/adaLN modulation is constant) and from avoiding differential-equation solver overhead.}}

\textbf{Energy Formulations.} We evaluate two EqM-E variants, the dot product and the $L_2$ norm, on the EqM-B/4 model. Since $L_2$ norm measures gradient norm and is intuitively consistent with the EqM objective, we also report $L_2$ norm results when finetuned from a pretrained EqM model. From \cref{energy}, we find that both energy formulations degrade performance when trained from scratch. We attribute this to the inherent difficulty of optimization, since training EqM-E would require calculating the Hessian and is numerically unstable. We also observe that finetuning an EqM-E $L_2$ model from a pretrained EqM model closes part of the gap to the implicit baseline and exceeds the performance of both from-scratch results. {We therefore view EqM-E primarily as a diagnostic that provides an exactly conservative reference field, rather than as a practical alternative to the implicit EqM model.}

\textbf{Empirical Conservativity of the Learned Field.}
Our EBM interpretation requires $f$ to be a conservative vector field. To verify whether this holds in practice, {we do not materialize the full high-dimensional Jacobian. Instead, we use forward-mode Jacobian-vector products with 256 probe pairs and 64 points per noise level} and report Jacobian asymmetry and curl diagnostics. A perfectly conservative field would give zero on both metrics. As shown in \cref{tab:conserv}, the raw Jacobian asymmetry remains below $4\!\times\!10^{-4}$ and the curl magnitude below $3\!\times\!10^{-3}$. Thus, EqM is approximately conservative and supports an approximate implicit-energy interpretation.

\begin{figure}[t]
\begin{minipage}[b]{0.53\textwidth}
\centering
\small
\begin{tabular}{ccccccc}
    \toprule
    noise level $\gamma$ & scale & 0.00 & 0.25 & 0.50 & 0.75 & 1.00 \\
    \midrule
    \makecell{$J$ asymmetry} & $\times 10^{-4}$ & 2.99 & 3.59 & 3.63 & 2.60 & 3.00 \\
    \makecell{curl magnitude} & $\times 10^{-3}$   & 2.4  & 2.9  & 2.9  & 2.1  & 2.4  \\
    \bottomrule
\end{tabular}
\vspace{-5pt}
  \captionof{table}{\textbf{Conservativity of the Learned Field.} Both Jacobian asymmetry
  and curl magnitude of the trained EqM model are close to 0 across all noise levels, indicating that the EqM field is approximately curl-free.}
  \label{tab:conserv}
\end{minipage}\hfill
\begin{minipage}[b]{0.44\textwidth}
\begin{center}
\small
\begin{tabular}{ccccc}
\toprule
 model & EqM & EqM-E & EqM-E  & EqM-E \\
 \midrule
 objective & - & dot & $L_2$ & finetuned \\
FID & 57.54 & 73.40 & 75.53 & 63.93 \\
\bottomrule
\end{tabular}
\vspace{-5pt}
\captionof{table}{\textbf{EqM-E Variants.} Finetuned EqM-E closes the gap to the baseline and outperforms other energy objectives, suggesting that EqM is inherently consistent with energy training.}
\label{energy}
\end{center}
\end{minipage}
\vspace{-5pt}
\end{figure}

\begin{figure}[t]
\begin{minipage}[b]{0.47\textwidth}
\vspace{0pt}    
\begin{center}
\small
\begin{tabular}{lcccc}
\toprule
model & SVHN & Textures  & constant & avg.\\
\midrule
\makecell{Pixel-\\CNN++}  & 0.32 & 0.33 & 0.71 & 0.45 \\
GLOW & 0.24 & 0.27 &  -  & 0.26\\
IGEBM &\textbf{0.63} &0.48 & 0.39 & 0.50\\
\textbf{EqM} & 0.55 &\textbf{0.49} & \textbf{1.00} & \textbf{0.68}\\
\bottomrule
\end{tabular}
\vspace{-5pt}
\captionof{table}{\textbf{OOD Detection.} EqM achieves reasonable AUROC under all tested OOD datasets and has the best average result among all tested models. }
\label{ood}
\end{center}
\end{minipage}\hfill
\begin{minipage}[b]{0.5\textwidth}
  \centering
  \includegraphics[scale=0.193]{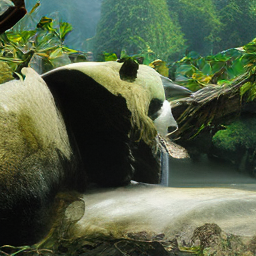}\hh
  \includegraphics[scale=0.193]{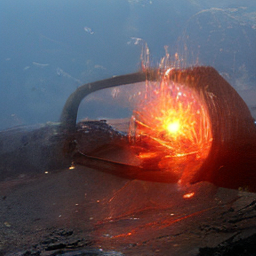}\hh
  \includegraphics[scale=0.193]{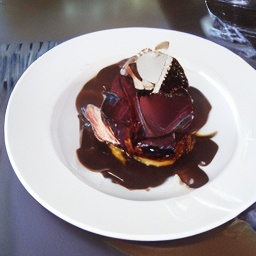}\hh
  \includegraphics[scale=0.193]{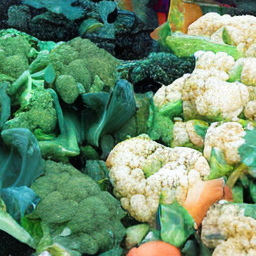}\\
\vspace{-5pt}
  \captionof{figure}{\textbf{Image Composition.} {We present compositional samples from EqM-XL/2 each using two ImageNet labels: panda and valley (leftmost), car mirror and volcano (second left), ice cream and chocolate syrup (second right), and broccoli and cauliflower (rightmost).} }
  \label{comp}
\end{minipage}\hfill
\vspace{-15pt}
\end{figure}

\subsection{Additional Properties}
\label{uni}
\begin{wrapfigure}{r}{0.4\textwidth}
\begin{minipage}[b]{1\linewidth}
  \centering
  \vspace{-40pt}
    \includegraphics[scale=0.35]{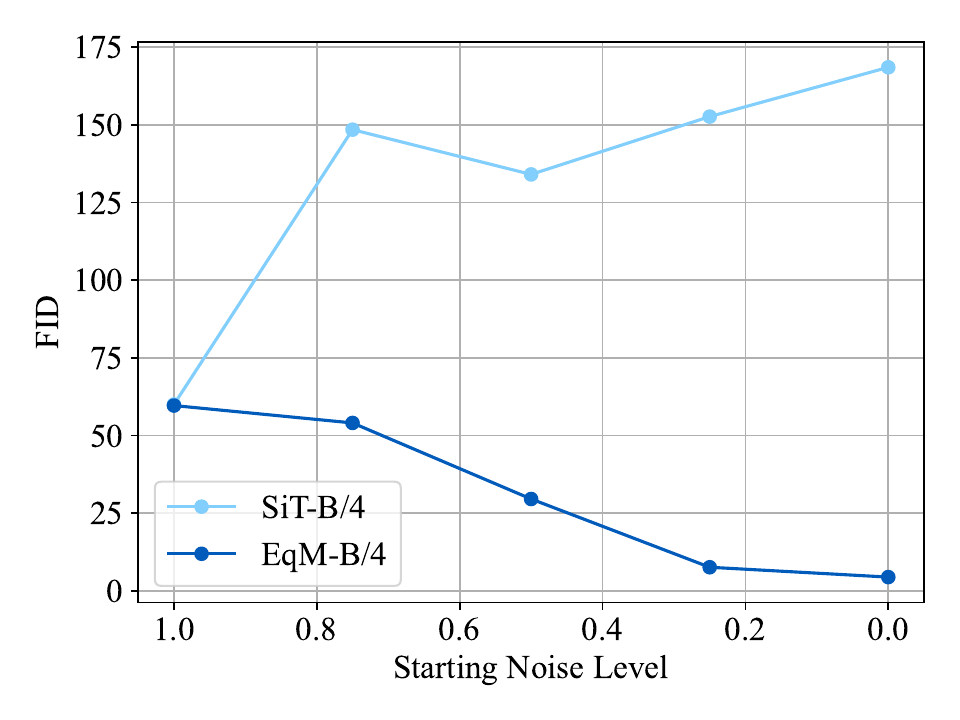}
\vspace{-15pt}
  \captionof{figure}{\textbf{Partially Noised Image Denoising.} EqM's generation quality improves with less noise while an FM baseline without the noise level drops. }
  \label{partial}
\end{minipage}
\vspace{-25pt}
\end{wrapfigure}
In this subsection, we investigate additional properties of Equilibrium Matching that are {not naturally supported by standard diffusion/flow formulations}.

\textbf{Partially Noised Image Denoising.} By learning equilibrium dynamics, EqM can directly start with and denoise a partially noised image. Existing diffusion/flow models require an explicit noise level as input to process such images, but EqM does not have this limitation. We evaluate EqM's generation quality from partially noised inputs using noised samples from the ImageNet validation set. As shown in \cref{partial}, EqM behaves very differently from Flow Matching {when the latter is not given the noise level}. EqM-B/4's FID improves significantly when fed less noisy samples, whereas the Flow-based SiT-B/4 cannot handle partially noised images as raw input and its generation quality drops quickly when not starting from pure noise. {An oracle FM/SiT baseline given the true noise level is reported in \cref{rebuttalcontrols}. EqM matches this oracle without requiring the noise level as input.}

\textbf{Out-of-Distribution Detection.} {An additional property} of the EqM model is its ability to perform out-of-distribution (OOD) detection using the energy value. In-distribution (ID) samples typically have lower energies than OOD samples. To this end, we use our dot product variant of the EqM-E-B/4 model and perform OOD detection with CIFAR-10 as ID. We use PixelCNN++ \citep{salimans2017pixelcnn++}, GLOW \citep{kingma2018glow}, and IGEBM \citep{du2019implicit} as baselines and adopt the numbers reported by \cite{yoon2023energy}. We report the area under the ROC curve (AUROC) in \cref{ood}. Compared with these baselines, EqM provides reasonable OOD detection across all tested datasets and achieves the best overall performance, {which is consistent with the implicit energy field interpretation}.

\textbf{Composition.} EqM also naturally supports the composition of multiple models by adding energy landscapes together (corresponding to adding the gradients of each model). We test composition by combining models conditioned on different ImageNet class labels. We use our EqM-XL/2 model with GD sampler and add two conditional gradients together as the update gradient at each sampling step. We present the generation results in \cref{comp}, where we observe semantic composition in produced images. We also report quantitative measurement of compositionality in \cref{compacc}. This is similar to the composability of EBMs \citep{du2020compositional}, while the composition of diffusion is significantly more complex to accurately implement \citep{du2023reduce}.

\section{Related Work}
\textbf{Diffusion, Flow Matching, and Energy-Based Models.} Diffusion models \citep{sohl,ddpm,ddim,nichol2021improved,dhariwal2021diffusion,edm} generate images from pure noise through noise-conditioned noising and denoising steps, with sampling typically formulated as integrating a predicted velocity along the noise schedule \citep{ddpm,song2020score,lu2022dpm,edm}. Flow Matching \citep{fm,recflow,albergo2023building} replaces the noise schedule with a linear interpolation between noise and data. EBMs \citep{hinton2002training,lecun2006tutorial,xie2016theory,du2019implicit,du2020improved,nijkamp2020anatomy,gao2020learning} learn an energy landscape representing the unnormalized log-density. Their equilibrium structure makes them versatile across modalities and downstream tasks \citep{du2019implicit,grathwohl2019your}, but they usually suffer from training instabilities \citep{carreira2005contrastive,song2018learning,gutmann2010noise} and are hard to scale \citep{du2020improved}.

\textbf{Hybrid Flow/EBM Methods.} Several prior works combine diffusion and energy-based training. \citet{sun2025noise} remove noise conditioning from diffusion models without changing the objective; Energy Matching \citep{balcerak2025energy} uses a two-stage scheme that first trains a flow objective and then refines with Langevin dynamics near the manifold. Along with VAPO \citep{yue2025vapo} and Action Matching \citep{neklyudov2023action}, these methods all retain time- or noise-conditioned objectives over time-dependent trajectories or actions. EqM differs by using a single objective that learns a stationary equilibrium gradient field describing the dynamics both near and away from data. We compare EqM against EBM-style methods in \cref{cifar}.

\section{Conclusion}
  
We propose Equilibrium Matching, a generative modeling framework that learns equilibrium dynamics in a simple and effective way. {EqM achieves competitive generation quality while providing an optimization-based perspective that connects flow and energy-based modeling.} Our method is easy to train, achieves strong generation quality, and provides {an approximately conservative implicit field (and an exact explicit-energy variant)} that supports a wide range of sampling methods. We hope that the equilibrium dynamics learned by EqM will inspire more flexible and scalable learning and inference algorithms in the future.

\section*{Acknowledgments}
We thank the Harvard Kempner Institute for granting us access to compute resources. We thank Michael Albergo, Hansen Lillemark, and Haldun Balim for helpful discussions. We thank Boyuan Chen for help on distributed training.
{
    \bibliographystyle{plainnat}
    \bibliography{ref}  \label{refs}
}


\newpage
\appendix
\section*{Appendix}
In \cref{expset}, we show the detailed experimental settings. In \cref{addexp}, we present additional experiments including CIFAR-10 experiments, {additional evaluation metrics}, sampling dynamics visualization, additional adaptive and stochastic optimizers, and quantitative compositionality measurements. In \cref{derivs}, we show the derivations of the statements made. In \cref{odeeqv}, we demonstrate the relationship between ODE-based integration sampler and our gradient descent sampler.
\section{Experimental Setting}
\label{expset}

\subsection{Training Setting}

We present the training setting of our Equilibrium Matching model in \cref{tab:combined_configs}.

\subsection{Inference Setting}
\label{inference}
We present the majority of our sampler settings in \cref{tab:combined_configs}. More specifically, in \cref{imagen}, we adopt the SiT results reported by \citet{sit} and \citet{wang2025diffuse}, and use our own NAG-GD sampler for EqM results. In \cref{opt}, \cref{ablation}, and \cref{uni}, we adopt the Euler sampler for SiT experiments and the vanilla GD sampler for EqM experiments.

\section{Additional Experiments}
\label{addexp}
\subsection{CIFAR-10 Experiments}  

We also evaluate our approach on non-transformer architectures on the CIFAR-10 dataset \citep{cifar}, where the commonly used network architecture is U-Net \citep{ronneberger2015u}.
The experiments are based on the publicly available code of Flow Matching \citep{lipman2024flow}.
{  We use the same hyperparameters as the original codebase, with the only difference being that we do not use EDM scheduling or skewed timesteps, which are tricks designed specifically for flow matching training \citep{edm}. \cref{cifar} presents the FID results. We used the truncated decay function with $a=0.4, \lambda=2.0$ for our EqM model. Equilibrium Matching outperforms Flow Matching baseline, similar to our observations on SiT. This demonstrates that EqM is a general generative modeling approach applicable across different datasets and model backbones.}

{   Since it has been hard to directly train EBMs on ImageNet and most existing literature only reports CIFAR-10 numbers, we compare other related methods' performance on CIFAR-10 in \cref{cifar}, including IGEBM \citep{du2019implicit, yang2023towards}, Energy-based U-net \citep{salimans2021should}, and Energy Matching \citep{balcerak2025energy}. EqM outperforms all listed methods, making it a promising approach for generative modeling. {As an additional cross-dataset/backbone check, on Fashion-MNIST with a U-Net and 600 epochs, EqM obtains FID-10k $7.48$ versus $11.32$ for FM under the same reference implementation.}}
\subsection{Additional Metrics}
\label{rebuttalcontrols}

We report additional evaluation metrics, including sFID and Inception Score (IS), in \cref{othermetrics}. Equilibrium Matching achieves competitive sFID and IS compared with other generative methods. We further evaluate distributional quality, adaptive stopping, and partially noised oracle controls below.

\paragraph{Distributional metrics.}
Using the identical XL/2 protocol, EqM improves or matches SiT on several complementary distributional metrics, as summarized in \cref{tab:distributional_metrics}. In DINOv2 feature space, EqM also achieves lower Fr\'echet distance and higher precision and coverage. These results complement FID rather than establish a distribution-level theorem.

\begin{table}[t]
    \centering

    \begin{tabular}{lcc}
        \toprule
        {metric} & {EqM} & {SiT} \\
        \midrule
        Precision ($k=3$) & 0.806 & 0.802 \\
        Recall ($k=3$) & 0.609 & 0.597 \\
        Density ($k=5$) & 1.130 & 1.143 \\
        Coverage ($k=5$) & 0.9997 & 0.9997 \\
        Mean per-class coverage & 0.943 (0.068) & 0.943 (0.070) \\
        \midrule
        DINOv2 Fr\'echet distance & 101.3 & 117.3 \\
        DINOv2 precision & 0.640 & 0.606 \\
        DINOv2 coverage & 0.760 & 0.714 \\
        \bottomrule
    \end{tabular}
        \caption{\textbf{EqM achieves competitive additional metrics.} Distributional metrics for EqM and SiT under the identical XL/2 protocol. Parentheses denote the standard deviation of per-class coverage.}
    \label{tab:distributional_metrics}
\end{table}

\paragraph{Adaptive quality--compute curve.}
For EqM-B/2, adaptive stopping produces the quality--compute trade-off shown in \cref{tab:adaptive_stopping}. A fixed 250-step EqM baseline obtains an FID-10k of 36.4. These results support a principled per-sample stopping criterion rather than a claim that adaptive stopping uniformly improves FID at every compute budget.

\begin{table}[t]
    \centering

    \begin{tabular}{cc}
        \toprule
        {mean NFE} & {FID-10k} \\
        \midrule
        657 & 36.0 \\
        241 & 36.5 \\
        168 & 38.1 \\
        141 & 39.6 \\
        \midrule
        250 (fixed) & 36.4 \\
        \bottomrule
    \end{tabular}
        \caption{\textbf{Quality--compute trade-off for adaptive stopping on EqM-B/2.}}
    \label{tab:adaptive_stopping}
\end{table}

\paragraph{Partially noised oracle control.}
Finally, we evaluate partially noised inputs at several starting noise levels. As shown in \cref{tab:oracle_control}, EqM closely matches an oracle SiT model that is explicitly given the true noise level, while substantially outperforming SiT without this input. Thus, EqM approximately recovers oracle-level performance without requiring the noise level as an input.

\begin{table}[t]
    \centering
    
    \begin{tabular}{lccc}
        \toprule
        & \multicolumn{3}{c}{{starting noise level}} \\
        \cmidrule(lr){2-4}
        {method} & {0.8} & {0.4} & {0.0} \\
        \midrule
        EqM & 59.2 & 20.2 & 6.2 \\
        SiT w/o noise-level input & 125.0 & 164.2 & 138.1 \\
        Oracle SiT & 58.2 & 20.2 & 5.4 \\
        \bottomrule
    \end{tabular}
    \caption{\textbf{EqM matches oracle FM without time input.} FID-10k for partially noised inputs. Oracle SiT is provided the true starting noise level, whereas EqM and standard SiT are not.}
    \label{tab:oracle_control}
\end{table}

\subsection{Energy Evolution}

\begin{figure}
    \centering
    \includegraphics[width=0.5\linewidth]{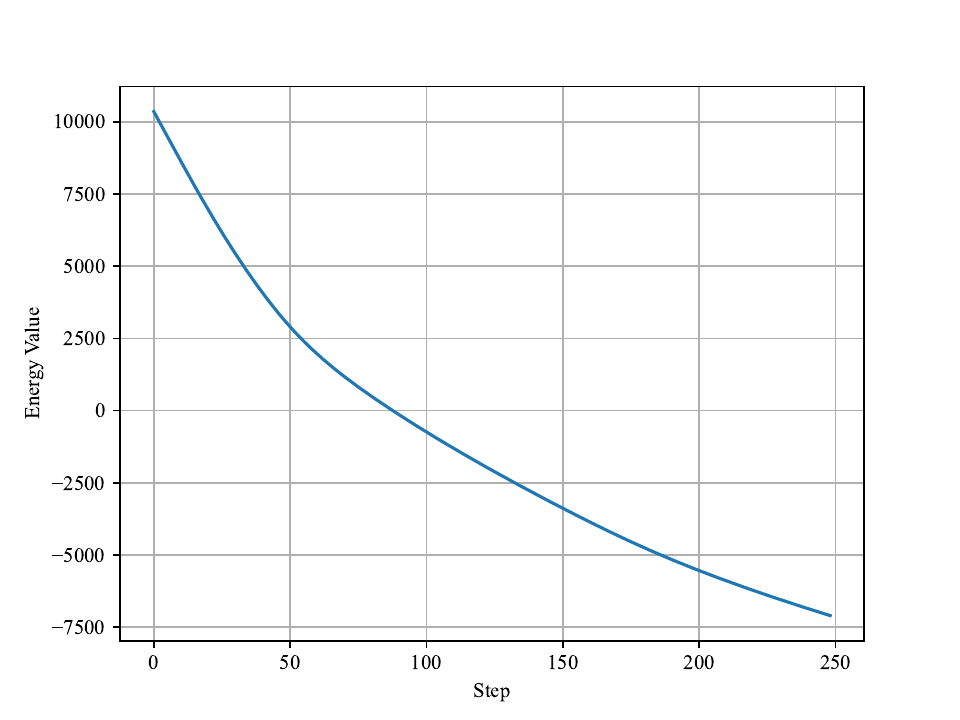}
    \caption{{  \textbf{Energy Evolution During Sampling.} We report the energy curve during sampling on EqM-E. The energy value decreases smoothly and consistently, matching our expectation.}}
    \label{energyevol}
\end{figure}

{   We include an energy evolution plot over the sampling process for our EqM-E model in \cref{energyevol}. We used the B/4 model with dot product parameterization and report the average energy curve over 64 samples. We see that the energy value consistently and smoothly decreases over the course of sampling, which is consistent with the expected behavior of an EBM.}

\subsection{Adaptive and Stochastic Optimizers}
\label{moreoptzer}
\paragraph{Adaptive optimizer.}
{  We also attempt to directly adopt existing adaptive optimizers for EqM sampling. In particular, we use Adam without momentum ($\beta_1=\beta_2=0$). We find that Adam performs decently well on EqM-B/2, as shown in \cref{opts}. This opens exciting opportunities for future research by demonstrating that reasonable samples can still be produced when the gradients are normalized.}

\begin{table}
    \centering
    \vspace{-0.5em}
    \begin{tabular}{l c}
        \toprule
        setting & {top-5 accuracy} \\
        \midrule
        single-class control & $0.2\%$ \\
        EqM composed samples & $1.8\%$ \\
        \bottomrule
    \end{tabular}
    \caption{\textbf{EqM composes classes with higher accuracy.} To measure composition, we report ResNet-50 top-5 classification on both labels of EqM composed image and compare against a single-class control paired with a random second class. EqM's composed samples provide much stronger signal to classification on both classes.}
    \label{fig:compacc}
\end{table}

\paragraph{Stochastic sampling.}
We additionally test whether the learned EqM field is compatible with stochastic sampling. On EqM-B/2, NAG-GD slightly improves both FID-10k and recall over deterministic gradient descent, while annealed Langevin sampling attains the same recall with a modest FID increase (\cref{tab:stochastic_sampling}).

\begin{table}[t]
    \centering
    
    \begin{tabular}{lcc}
        \toprule
        {sampler} & {FID-10k} & {Recall} \\
        \midrule
        GD & 20.42 & 0.627 \\
        NAG-GD & 20.34 & 0.634 \\
        Annealed Langevin & 21.11 & 0.634 \\
        \bottomrule
    \end{tabular}
    \caption{\textbf{EqM supports stochastic sampling.} Stochastic sampling results for EqM-B/2. Annealed Langevin sampling uses a cosine schedule with $T_0=0.01$.}
    \label{tab:stochastic_sampling}
\end{table}
\subsection{Measuring Compositionality}
    \label{compacc}
{To measure compositionality of EqM models in a quantitative way, we report ResNet-50 top-5 classification on both labels of EqM composed image and compare against a single-class control paired with a random second class. As reported in \cref{fig:compacc}, across 100 random class pairs with 20 samples each, $1.8\%$ of composed samples contain both parent classes in a ResNet-50 top-5 prediction, compared with $0.2\%$ for a single-class control paired with a random second class. EqM therefore composes class labels in a semantically meaningful way detectable by classifiers.} 
\section{Derivations}
\label{derivs}
\subsection{Learned Gradient at Ground-Truth Samples}
\label{st1der}

\begin{statement}[Population-Optimal Field at Training Samples]
{Let $f^*$ be the population-optimal EqM field, let $x^{(i)}\in\mathcal X\subset\mathbb R^d$, and suppose $c(1)=0$ with $c(\gamma)\le C(1-\gamma)$ near $1$. For the continuous version of the conditional-mean field and $d\ge3$, $f^*(x^{(i)})=0$.}
\end{statement}
\begin{table}[t]
  \centering
  \begin{tabular}{lcccc}
    \toprule
    model & S/2 & B/2 & L/2 & XL/2 \\
    \midrule
    \multicolumn{5}{l}{\bfseries model configurations} \\
    params (M)           &  33  & 130  & 458  & 675   \\
    depth                &  12  &  12  &  24  &  28        \\
    hidden dim           & 384  & 768  &1024  &1152       \\
    patch size           &  2   &  2   &  2   &  2     \\
    heads                &  6   & 12   & 16   & 16  \\
    \midrule
    \multicolumn{5}{l}{\bfseries ImageNet training configurations} \\
    epochs & 80 & 80 & 80 & 80 - 1400 \\
    batch size           & \multicolumn{4}{c}{256}   \\
    optimizer            & \multicolumn{4}{c}{AdamW}  \\
    optimizer  $\beta_1$          & \multicolumn{4}{c}{0.9}  \\
    optimizer  $\beta_2$          & \multicolumn{4}{c}{0.999}  \\
    weight decay         & \multicolumn{4}{c}{0.0}     \\
    learning rate (lr)   & \multicolumn{4}{c}{$1\times10^{-4}$}     \\
    lr schedule          & \multicolumn{4}{c}{constant}      \\
    lr warmup            & \multicolumn{4}{c}{none}      \\
    \midrule
    \multicolumn{5}{l}{\bfseries gradient field hparams} \\
    choice of $c(\gamma)$&\multicolumn{4}{c}{truncated decay}\\
    $c(\gamma)$ multiplier $\lambda$ &      \multicolumn{4}{c}{4.0}   \\
    $a$ &      \multicolumn{4}{c}{0.8}   \\
    $b$ &      \multicolumn{4}{c}{-}   \\
    \midrule
    \multicolumn{5}{l}{\bfseries integration sampler configurations} \\
    sampler & dopri5 & dopri5 & dopri5 & Euler \\
    NFE & \multicolumn{4}{c}{250} \\
    $\eta$ & - & - & - & 0.0017 \\
    \midrule
    \multicolumn{5}{l}{\bfseries gradient sampler configurations} \\
    sampler & \multicolumn{4}{c}{GD/NAG-GD} \\
    steps & \multicolumn{4}{c}{250} \\
    $\eta$ & 0.003 & 0.003 & 0.003 & 0.0017 \\
    $\mu$ & 0.35 & 0.35 & 0.35 & 0.3 \\
    \bottomrule
  \end{tabular}
    \caption{\textbf{Equilibrium Matching Configurations.}}
    \label{tab:combined_configs}
\end{table}
\begin{proof}[Derivation of Statement~\ref{st1}]
{
At the population optimum,
\[
f^*(z)=\mathbb E[(\epsilon-x)c(\gamma)\mid x_\gamma=z].
\]
Because conditioning on an exact point is a zero-measure event, we make the argument through shrinking balls. Let $u=1-\gamma$ and condition on $B_r(x^{(i)})$. On the branch generated by $x=x^{(i)}$, the event implies $\|u(\epsilon-x^{(i)})\|\le r$. Since $c(1-u)\le Cu$,
\[
\|(\epsilon-x^{(i)})c(1-u)\|\le Cr.
\]
This branch has probability $\Theta(r)$ as $r\to0$. For every other training point, the corrupted sample has a smooth $d$-dimensional density around $x^{(i)}$, so its probability of entering the ball is $O(r^d)$ and its conditional target first moment is bounded. Therefore
\[
\left\|\mathbb E[(\epsilon-x)c(\gamma)\mid x_\gamma\in B_r(x^{(i)})]\right\|
=O\!\left(C(r+r^{d-1})\right)\xrightarrow[r\to0]{}0.
\]
The continuous version of the population-optimal field hence satisfies $f^*(x^{(i)})=0$. This avoids assigning positive probability to the event $\gamma=1$.
}
\end{proof}

\subsection{Local Attraction and Non-Data Equilibria}
\label{st11der}

\begin{statement}[Scope of Population-Optimal Equilibria]
{For linear decay $c(\gamma)=\lambda(1-\gamma)$, the population-optimal field is
\[
f^*(z)=\lambda\bigl(z-\mathbb E[x\mid x_\gamma=z]\bigr).
\]
Hence every zero lies in $\operatorname{conv}(\mathcal X)$, and an isolated training point is a locally attracting critical point. For non-data zeros we make the resolvability assumption below; no claim is made for arbitrary non-resolvable zeros.}
\end{statement}

\begin{proof}[Derivation and scope of Statement~\ref{st11}]
{
For $u=1-\gamma$, linear decay makes the target
\[
(\epsilon-x)c(\gamma)=\lambda u(\epsilon-x)=\lambda(x_\gamma-x),
\]
which gives the posterior-mean form above. At a zero $\hat x$, $\hat x=\mathbb E[x\mid x_\gamma=\hat x]$, so $\hat x$ lies in the convex hull of the training set. Near an isolated training point the posterior concentrates on that point, yielding a positive local Jacobian for $f^*$ and therefore a contractive gradient-descent update for sufficiently small step size.

\textbf{Resolvability assumption.} A non-data zero $\hat x$ is $(k,\kappa)$-resolvable if, for some constant $c_0>0$, all but an $e^{-c_0 d}$ fraction of the posterior mass lies on $k$ training points with distances $D_j=\|x^{(j)}-\hat x\|$ satisfying $\max_j D_j/\min_j D_j\le\kappa$. 

Let $v$ be the top spread direction of these supporting components and $a_j=v^T(x^{(j)}-\hat x)$. The balance condition gives $\sum_j w_j a_j=0$. A per-branch Laplace evaluation gives
\[
\mathbb E[(1-u)/u^2\mid j]=\frac{d}{D_j^2}\bigl(1+O(1/d)\bigr),
\]
and the spread bound gives $\sum_j w_j a_j^2\ge (\sum_j w_jD_j^2)/(k-1)$. Consequently the directional Jacobian is bounded, up to the cross term from unequal $D_j$, by
\[
v^TJ_{f^*}(\hat x)v\;\lesssim\;\lambda\!\left(1-\frac{d}{(k-1)\kappa^2}\right)+O(\kappa-1).
\]
For equidistant supporting components ($\kappa=1$) the cross term vanishes, so any resolvable non-data zero with $k-1<d$ has a negative-curvature/expansive direction and is not a stable minimum; the two-component case is $k=2$. For $\kappa$ sufficiently close to one, the same conclusion holds by perturbation. Thus the theory rules out stable non-data equilibria only under this explicit resolvability condition. 
}
\end{proof}

\begin{figure}[t]
\begin{minipage}[b]{1\linewidth}
\begin{center}
\begin{tabular}{cccccc}
\toprule
model &FM & IGEBM & \makecell{Energy-based\\ U-net} & \makecell{Energy\\Matching} & EqM\\
\midrule
FID & 3.70 & 37.9 & 6.8 & 3.34 & \textbf{3.32} \\
\bottomrule
\end{tabular}
\captionof{table}{{  \textbf{Image Generation on CIFAR-10.} We report FID on CIFAR-10 class-conditional image generation with U-Net. EqM improves upon Flow Matching and energy-based methods on CIFAR-10 generation.}}
\label{cifar}
\end{center}
\end{minipage}

\vspace{2em}
\begin{minipage}[b]{1\linewidth}
    \centering
    \begin{tabular}{lcccc}
\toprule
model & method &  FID $\downarrow$ & sFID $\downarrow$& IS $\uparrow$\\
\midrule
StyleGAN-XL &GAN& 2.30&  \textbf{4.02}& 265.1\\
VDM++ & Diffusion  & 2.12 & - &  267.7\\
DiT-XL/2 & Diffusion  & 2.27 & 4.60 & \textbf{278.2}\\
SiT-XL/2 &  FM  & 2.06 & 4.49 & 277.5 \\
\textbf{EqM-XL/2} & \textbf{EqM}  & \textbf{1.90} &4.54 &275.7 \\
\bottomrule
\end{tabular}
    \captionof{table}{\textbf{Class-conditional Generation on ImageNet 256$\times$256 with Additional Metrics.} Equilibrium Matching achieves the best FID and relatively good sFID and IS.}
    \label{othermetrics}
\end{minipage}
\end{figure}

\subsection{Convergence of Gradient-Based Sampling}
\label{st2der}

\begin{statement}[Convergence of Gradient-Based Sampling on an Explicit Energy]
    {Let $E$ be an explicit EqM-E energy and $f=\nabla E$.} Suppose $E$ is $L$-smooth and bounded below by $E(x)\geq E_\text{inf}$. Then, gradient descent with step size $\eta \in \bigl(0,\tfrac{1}{L}\bigr]$ satisfies:
    \[
    \min_{0\leq k < K}\|f(x_k)\|^2 \leq \frac{2\bigl(E(x_0)-E_\text{inf}\bigr)}{\eta\,K},
    \]
    where $x_k$ is the iterate after $k$ steps and $K$ is the total number of optimization steps performed. {For implicit EqM, this is an approximate energy-descent interpretation.}
\end{statement}

\begin{proof}[Derivation of Statement~\ref{st2}]
\renewcommand{\theequation}{\arabic{equation}}
\setcounter{equation}{0}

By $L$-smoothness of $E$, for any $x,y\in\mathbb R^d$,
\[
E(y)\,\le\,E(x) + \nabla E(x)^\top(y-x) + \frac{L}{2}\,\|y-x\|^2.
\]

Take $y = x_{k+1} = x_k - \eta\,f(x_k)$ and recall \rev{$\nabla E(x_k) = f(x_k)$}. Then
\begin{align*}
E(x_{k+1})
&\le
E(x_k)
- \eta\,f(x_k)^\top f(x_k)
+ \tfrac{L}{2}\,\eta^2\,\|f(x_k)\|^2
\\
&=
E(x_k)
- \eta\Bigl(1 - \tfrac{L\eta}{2}\Bigr)\|f(x_k)\|^2
\;\;\le\;\;
E(x_k) - \tfrac{\eta}{2}\|f(x_k)\|^2,
\end{align*}
where the last inequality holds because $\eta\le 1/L\implies1 - \tfrac{L\eta}{2}\ge\tfrac12$.

Summing from $k=0$ to $k=K-1$ gives
\[
E(x_K) - E(x_0)
\;\le\;
-\tfrac{\eta}{2}\sum_{k=0}^{K-1}\|f(x_k)\|^2
\;\le\;
-\tfrac{\eta}{2}\,K \,\min_{0\le k<K}\|f(x_k)\|^2.
\]
Since $E(x_K)\ge E_\mathrm{inf}$, rearrange to obtain
\[
\min_{0\le k<K}\|f(x_k)\|^2
\;\le\;
\frac{2\bigl(E(x_0)-E_\mathrm{inf}\bigr)}{\eta\,K},
\]
as required.
\end{proof}

\section{Relation between Integration-Based and Optimization-Based Sampling}
\label{odeeqv}

\subsection{ODE Sampling as a Special Case of Optimization}
Consider the ODE
\[
\dot{x} = v(x),
\]
and its explicit (forward) Euler discretization with a uniform time grid on $[0,1]$:
\[
x_{k+1} = x_k + h\, v(x_k), \qquad h=\tfrac{1}{N},\; k=0,\dots,N-1.
\]
When the velocity field is conservative with potential $E$, i.e., $v(x) = -\nabla E(x)$, the Euler step becomes:
\[
x_{k+1} = x_k - h\, \nabla E(x_k),
\]
which is exactly a gradient-descent update with step size $\eta=h=\tfrac{1}{N}$. Hence, with $N$ steps on a unit time horizon, the gradient-descent sampler with $\eta=\tfrac{1}{N}$ coincides with the explicit Euler ODE sampler.

\subsection{General Integration Samplers}
The equivalence above suggests a broader correspondence: integration-based samplers can be interpreted as optimization-based methods by viewing the velocity as a descent direction. In particular, when $v(x)=-\nabla E(x)$, any time integrator induces an optimization update rule.

Practically, ODE samplers in generative modeling are often implemented on a uniform grid with step size $h=\tfrac{1}{N}$. Under the optimization view, we are not bound to this constraint. We can adopt adaptive step sizes while retaining the same underlying direction field. This perspective enables a direct adaptation of existing integration-based samplers within Equilibrium Matching.

\begin{table}[]
    \centering
    \begin{tabular}{ccccc}
    \toprule
      sampler  & GD & NAG-GD & Adam \\
      \midrule
        FID & 32.85 & 32.97 & 36.35\\
    \bottomrule
    \end{tabular}
    \caption{  \textbf{EqM Sampling with Adam.} We find that Adam without momentum achieves decent generation quality on EqM-B/2. }
    \label{opts}
\end{table}


\newpage
\section*{NeurIPS Paper Checklist}

\begin{enumerate}

\item {\bf Claims}
    \item[] Question: Do the main claims made in the abstract and introduction accurately reflect the paper's contributions and scope?
    \item[] Answer: \answerYes{} 
    \item[] Justification: We conduct extensive experiments in the Experiments section to validate our claims.

\item {\bf Limitations}
    \item[] Question: Does the paper discuss the limitations of the work performed by the authors?
    \item[] Answer: \answerYes{} 
    \item[] Justification: We explicitly discuss the assumptions on high dimensionality and conservativeness in Section 4 Analysis and verify that these assumptions hold in practice in Section 5 Experiments.

\item {\bf Theory assumptions and proofs}
    \item[] Question: For each theoretical result, does the paper provide the full set of assumptions and a complete (and correct) proof?
    \item[] Answer: \answerYes{} 
    \item[] Justification: We provide the assumptions in Analysis and provide the full correct proof in Appendix.

    \item {\bf Experimental result reproducibility}
    \item[] Question: Does the paper fully disclose all the information needed to reproduce the main experimental results of the paper to the extent that it affects the main claims and/or conclusions of the paper (regardless of whether the code and data are provided or not)?
    \item[] Answer: \answerYes{} 
    \item[] Justification: We disclose our code to reproduce the main experiments and provide all details needed for experiments in Appendix.

\item {\bf Open access to data and code}
    \item[] Question: Does the paper provide open access to the data and code, with sufficient instructions to faithfully reproduce the main experimental results, as described in supplemental material?
    \item[] Answer: \answerYes{} 
    \item[] Justification: We provide a well documented codebase with readme detailing reproduction procedures.

\item {\bf Experimental setting/details}
    \item[] Question: Does the paper specify all the training and test details (e.g., data splits, hyperparameters, how they were chosen, type of optimizer) necessary to understand the results?
    \item[] Answer: \answerYes{} 
    \item[] Justification: We include all details in Appendix.

\item {\bf Experiment statistical significance}
    \item[] Question: Does the paper report error bars suitably and correctly defined or other appropriate information about the statistical significance of the experiments?
    \item[] Answer: \answerYes{} 
    \item[] Justification: We conduct experiments across at least 3 scales to verify statistically consistent improvement.

\item {\bf Experiments compute resources}
    \item[] Question: For each experiment, does the paper provide sufficient information on the computer resources (type of compute workers, memory, time of execution) needed to reproduce the experiments?
    \item[] Answer: \answerYes{} 
    \item[] Justification: We detail the compute instance we used for our wall-clock time experiments.
    
\item {\bf Code of ethics}
    \item[] Question: Does the research conducted in the paper conform, in every respect, with the NeurIPS Code of Ethics \url{https://neurips.cc/public/EthicsGuidelines}?
    \item[] Answer: \answerYes{} 
    \item[] Justification: We faithfully follow the NeurIPS code of ethics.

\item {\bf Broader impacts}
    \item[] Question: Does the paper discuss both potential positive societal impacts and negative societal impacts of the work performed?
    \item[] Answer: \answerNA{} 
    \item[] Justification: Our work focus on improving generative modeling method on image generation, which is a well established area and we strictly follow prior literature.
    
\item {\bf Safeguards}
    \item[] Question: Does the paper describe safeguards that have been put in place for responsible release of data or models that have a high risk for misuse (e.g., pre-trained language models, image generators, or scraped datasets)?
    \item[] Answer: \answerNA{} 
    \item[] Justification: We do not involve high risk for misuse.

\item {\bf Licenses for existing assets}
    \item[] Question: Are the creators or original owners of assets (e.g., code, data, models), used in the paper, properly credited and are the license and terms of use explicitly mentioned and properly respected?
    \item[] Answer: \answerYes{} 
    \item[] Justification: We credit all assets used in this paper.

\item {\bf New assets}
    \item[] Question: Are new assets introduced in the paper well documented and is the documentation provided alongside the assets?
    \item[] Answer: \answerYes{} 
    \item[] Justification: We provide sufficient documentation for our codebase.

\item {\bf Crowdsourcing and research with human subjects}
    \item[] Question: For crowdsourcing experiments and research with human subjects, does the paper include the full text of instructions given to participants and screenshots, if applicable, as well as details about compensation (if any)? 
    \item[] Answer: \answerNA{} 
    \item[] Justification: No crowdsourcing or human subjects are involved.

\item {\bf Institutional review board (IRB) approvals or equivalent for research with human subjects}
    \item[] Question: Does the paper describe potential risks incurred by study participants, whether such risks were disclosed to the subjects, and whether Institutional Review Board (IRB) approvals (or an equivalent approval/review based on the requirements of your country or institution) were obtained?
    \item[] Answer: \answerNA{} 
    \item[] Justification: The paper does not involve crowdsourcing nor research with human subjects.

\item {\bf Declaration of LLM usage}
    \item[] Question: Does the paper describe the usage of LLMs if it is an important, original, or non-standard component of the core methods in this research? Note that if the LLM is used only for writing, editing, or formatting purposes and does \emph{not} impact the core methodology, scientific rigor, or originality of the research, declaration is not required.
    \item[] Answer: \answerNA{} 
    \item[] Justification: The core method development in this research does not involve LLMs as any important, original, or non-standard components.

\end{enumerate}

\end{document}